\documentclass[sigconf]{acmart}
\AtBeginDocument{%
  }

\acmISBN{978-1-4503-XXXX-X/2018/06}

\usepackage{subfig}
\usepackage{algorithm}
\usepackage{algorithmic}
\usepackage{threeparttable}

\begin{document}

\title{Mamba Meets Scheduling: Learning to Solve Flexible Job Shop Scheduling with Efficient Sequence Modeling}


\author{Zhi Cao}
\email{cz20000927@mail.dlut.edu.cn}
\affiliation{%
  \institution{Dalian University of Technology}
 \city{Dalian}
 \country{China}
}

\author{Cong Zhang}
\email{cong.zhang92@gmail.com}
\affiliation{%
  \institution{Nanyang Technological University}
  \country{Singapore}}

\author{Yaoxin Wu}
\email{y.wu2@tue.nl}
\affiliation{%
  \institution{Eindhoven University of Technology}
  \city{Eindhoven}
  \country{the Netherlands}
}

\author{Yaqing Hou}
\email{houyq@dlut.edu.cn}
\affiliation{%
 \institution{Dalian University of Technology}
 \city{Dalian}
 \country{China}
 }

\author{Hongwei Ge}
\email{ hwge@dlut.edu.cn}
\affiliation{%
  \institution{Dalian University of Technology}
 \city{Dalian}
 \country{China}
  }





\begin{abstract}
The Flexible Job Shop Problem (FJSP) is a well-studied combinatorial optimization problem with extensive applications for manufacturing and production scheduling. It involves assigning jobs to various machines to optimize criteria, such as minimizing total completion time. Current learning-based methods in this domain often rely on localized feature extraction models, limiting their capacity to capture overarching dependencies spanning operations and machines. This paper introduces an innovative architecture that harnesses Mamba, a state-space model with linear computational complexity, to facilitate comprehensive sequence modeling tailored for FJSP. In contrast to prevalent graph-attention-based frameworks that are computationally intensive for FJSP, we show our model is more efficient. Specifically, the proposed model possesses an encoder and a decoder. The encoder incorporates a dual Mamba block to extract operation and machine features separately. Additionally, we introduce an efficient cross-attention decoder to learn interactive embeddings of operations and machines. Our experimental results demonstrate that our method achieves faster solving speed and surpasses the performance of state-of-the-art learning-based methods for FJSP across various benchmarks.
\end{abstract}




\maketitle

\section{Introduction}
The Flexible Job Shop Problem (FJSP) extends the classical Job Shop Scheduling Problem (JSSP) and is a significant combinatorial optimization challenge in computer science and operations research. It is prevalent in manufacturing, automotive, and aerospace \cite{brandimarte1993routing}. Unlike JSSP, FJSP allows for a more complex assignment of jobs, where each operation can be processed by any machine from a set of eligible machines, introducing additional flexibility and complexity in scheduling \cite{dauzere2024flexible}. FJSP aims to assign jobs to a set of heterogeneous machines while optimizing objectives such as minimizing the makespan (total completion time of all jobs), total flowtime, or tardiness. Due to its NP-hard nature, solving FJSP optimally is highly impractical, and therefore, practical solutions often depend on heuristic approaches \cite{fattahi2007mathematical} or approximate algorithms \cite{shahgholi2019heuristic, jansen2000approximation}. These methods aim to provide near-optimal solutions within a reasonable computational time, balancing quality and efficiency.

In the current research landscape, there has been a notable increase in investigation focused on integrating machine learning techniques to address the challenging JSSP and FJSP. Notably, JSSP has garnered significant attention and scrutiny within these efforts \cite{zhang2020learning,zhang2024deep,zhang2024learning,park2021learning,tassel2021reinforcement,iklassov2022learning}. Conversely, FJSP remains relatively unexplored, primarily due to the intricate complexity inherent in the problem. Current research focusing on learning to tackle FJSP often casts the problem as a sequential decision problem. They leverage deep reinforcement learning to tackle this intricate challenge. In their approaches, the states are typically represented by directed acyclic graphs (known as disjunctive graphs), over which the decisions (i.e., selecting a candidate operation to be assigned and scheduled to one of the compatible machines) are made. These graph-structured states are first mapped to latent space by extracting embeddings using deep neural nets, with graph neural networks as a fundamental component \cite{smit2024graph}. Due to the more powerful representation learning ability, an emerging trend utilizes the graph-attention mechanisms to replace graph convolutions for learning graph embeddings \cite{wang2024flexible}. 
Nevertheless, both models focus on extracting operation and machine embeddings within a localized neighbourhood surrounding the target, following the graph's topology. This approach may hinder comprehensive representation learning, leading to constrained performance outcomes.
Furthermore, the computational overhead associated with graph attention challenges the efficiency required for addressing the FJSP, a critical dimension of the method's overall performance. 
Henceforth, designing node connections in disjunctive graphs is always tricky and head-scratching for an effective trade-off between computational efficiency and performance \cite{wang2024flexible}.
Therefore, a natural question is, can we develop a neural architecture that can improve the performance based on learning the full sequence of operations (and machines) to bypass the need for elaborate graph-based state design and maintain faster-solving speed at the same time?

In this paper, we answer this question by proposing \textbf{\underline{M}}amba-\textbf{\underline{C}}ross\textbf{\underline{A}}ttention, a novel neural network with Mamba \cite{gu2023mamba} as the core, to enable efficient and comprehensive sequence modelling for FJSP. M-CA comprises an encoder and a decoder. The encoder extracts operation and machine features separately. In specific, the encoder integrates dual Mamba blocks to separately extract latent representations for operation-specific attributes (e.g., processing times, precedence constraints) and machine-specific states (e.g., workload, availability). This bifurcated design ensures specialized feature learning while mitigating interference between heterogeneous data modalities. The decoder then employs a lightweight cross-attention mechanism to fuse these representations, enabling the model to jointly reason about operation-machine assignments and scheduling sequences. By replacing computationally expensive graph-attention layers with Mamba’s linear-complexity structured state space models (SSMs), we demonstrate that M-CA achieves significant efficiency gains without sacrificing solution quality.

In summary, our contributions are as follows:
\begin{itemize}
    \item We extend the boundary of Mamba models to scheduling domains by showing that Mamba is effective and efficient in learning to solve FJSP. To our knowledge, this is the first attempt to apply Mamba to solving discrete combinatorial optimization problems in manufacturing scheduling.
    \item We proposed the Mamba-CrossAttention network to enable high-quality end-to-end solution generation for FJSP. It learns the interactive representation of operations and machines from their full sequence, breaking the neighbourhood restriction of existing graph-based approaches.
    \item We show through extensive experiments that our approach can provide near-optimal solutions for FJSP, achieving new state-of-the-art results among learning-based approaches with a faster solving speed.
\end{itemize}


\section{Related Work}
The rapid progress of artificial intelligence has sparked a new surge of interest in approaching manufacturing scheduling problems through a machine learning perspective, particularly leveraging deep learning techniques~\cite{dogan2021machine,smit2025graph,reijnen2023job}. For the Job Shop Scheduling Problem (JSSP), neural differentiable methods rooted in deep reinforcement learning (DRL) have emerged as the predominant machine learning paradigm. The prevailing neural approaches for JSSP predominantly favoured construction heuristics, which learn to extend partial solutions to complete ones sequentially. Notably, L2D~\cite{zhang2020learning} stands as a milestone work in this domain, where a GIN-based policy learns latent embeddings of partial solutions, depicted as disjunctive graphs and selects operations for assignment to respective machines at each construction step. Similar dispatching strategies can be observed in RL-GNN~\cite{park2021learning} and ScheduleNet~\cite{park2021schedulenet}, which introduce artificial machine nodes containing machine-progress information into the disjunctive graph to integrate machine status into decision-making. These augmented disjunctive graphs are treated as undirected, with a type-aware GNN model featuring two independent modules proposed for extracting machine and task node embeddings. Despite notable advancements over L2D, the performance remains suboptimal. DGERD~\cite{chen2022deep} follows a procedure akin to L2D but employs a Transformer-based embedding network~\cite{vaswani2017attention}. In a recent development, MatNet~\cite{NEURIPS2021_29539ed9} adopts an encoding-decoding framework for learning construction heuristics for flexible flow shop problems. However, its assumption of independent machine groups for operations at each stage is overly restrictive for JSSP. JSSenv~\cite{tassel2021reinforcement} introduces a meticulously designed and well-optimized simulator for JSSP, instantiated as an extension of the OpenAI gym environment suite~\cite{brockman2016openai}. Instead of utilizing disjunctive graphs, JSSenv models and represents partial schedule states using Gantt charts~\cite{jain1999deterministic}, and also introduces a DRL agent to learn to solve JSSP instances individually online. Nonetheless, its online nature, necessitating training for each instance, is less rewarding than offline-trained methodologies that can solve unseen problem instances once trained. In addition to construction heuristics, there are learning-to-improve methods that iteratively refine an initial solution to ones with better quality \cite{zhang2024deep,zhang2024learning}.

Unlike JSSP, which has been vastly studied, relatively little attention has been invested in FJSP. Song et al. \citep{song2023flexible} introduce a Deep Reinforcement Learning (DRL) based method to address FJSP by learning high-quality priority dispatching rules (PDRs) end-to-end. The approach combines operation selection and machine assignment into a unified decision-making process, utilizing a unique heterogeneous graph representation of scheduling states that captures complex relationships between operations and machines. DAN \cite{wang2024flexible} replaces the graph convolution kernel with the graph-attention mechanism. DAN consists of interconnected operation message attention blocks and machine message attention blocks, enabling the precise representation of complex relationships between the two. 
Despite the promising results, their methods require a tailored design of the graph-based state. The representation learning over these graphs is confined within a localized neighbourhood for each operation and machine, therefore heavily relying on message-passing to capture the holistic context of the problem, potentially limiting the overall performance, especially when utilizing shallow Graph Neural Networks (GNNs) or encountering over-smoothing issues with deeper GNN architectures \cite{rusch2023survey}.

\section{Preliminaries}

\subsection{Mamba Network}


\textbf{State Space Models.}
The State Space Models (SSM) maps a continuous input signal $x(t) \in \mathbb{R}$ to a corresponding output $y(t) \in \mathbb{R}$ through the state representation $h(t) \in \mathbb{R}^N$. This state space describes the evolution of the state over time and can be expressed using ordinary differential equations as follows:

\begin{equation}
	\begin{aligned}
		h^\prime(t) & =\boldsymbol{A}h(t)+\boldsymbol{B}x(t) \\
		y(t) & =\boldsymbol{C}h(t)+\boldsymbol{D}x(t)
	\end{aligned}
\end{equation}
where $h^\prime(t)=\frac{dh(t)}{dt}$, and $\boldsymbol{A}, \boldsymbol{B}, \boldsymbol{C}$, and $\boldsymbol{D}$ are learnable parameter matrices.

\noindent\textbf{Discretization.}
Due to the continuous nature of SSM, finding its analytical solution is difficult. Also to facilitate the solution of discrete sequence inputs, such as operation and machine sequences in scheduling, discretization methods are introduced. Through discretization, SSM is able to treat discrete sequences as samples of a continuous signal at fixed time intervals. The resulting discrete state space model can be expressed as follows:
\begin{equation}
	\begin{aligned}
		h_k & =\overline{\boldsymbol{A}} h_{k-1}+\overline{\boldsymbol{B}} x_k \\
		y_k & =\overline{\boldsymbol{C}} h_k+\overline{\boldsymbol{D}} x_k
	\end{aligned}
\end{equation}
where $h_k$ denotes the state vector at moment $k$ and $x_k$ denotes the input vector at moment $k$. The continuous time matrices $A$ and $B$ are transformed into discrete $\overline{A}$ and $\overline{B}$ matrices by applying some discretization techniques such as the ZOH (Zero Order Holding) method. In this case, $\overline{\boldsymbol{A}}=\exp (\Delta \boldsymbol{A}), \overline{\boldsymbol{B}}=(\Delta \boldsymbol{A})^{-1}(\exp (\Delta \boldsymbol{A})-\boldsymbol{I}) \cdot \Delta \boldsymbol{B}$.

\noindent\textbf{Selective Scan Mechanism.}
Mamba further introduces selective SSM by parameterizing the inputs to the SSM. This allows the model to process sequences selectively to focus on or ignore specific inputs. This selective mechanism allows the Mamba model to strike a balance between sequence feature extraction capability and computational efficiency, giving it the potential to outperform Transformer. In addition, Mamba has been designed with hardware-aware algorithms that enable parallel scan training.

\subsection{Flexible Job Shop Scheduling}
\begin{figure}[h]
\centering
\includegraphics[width=0.48\textwidth]{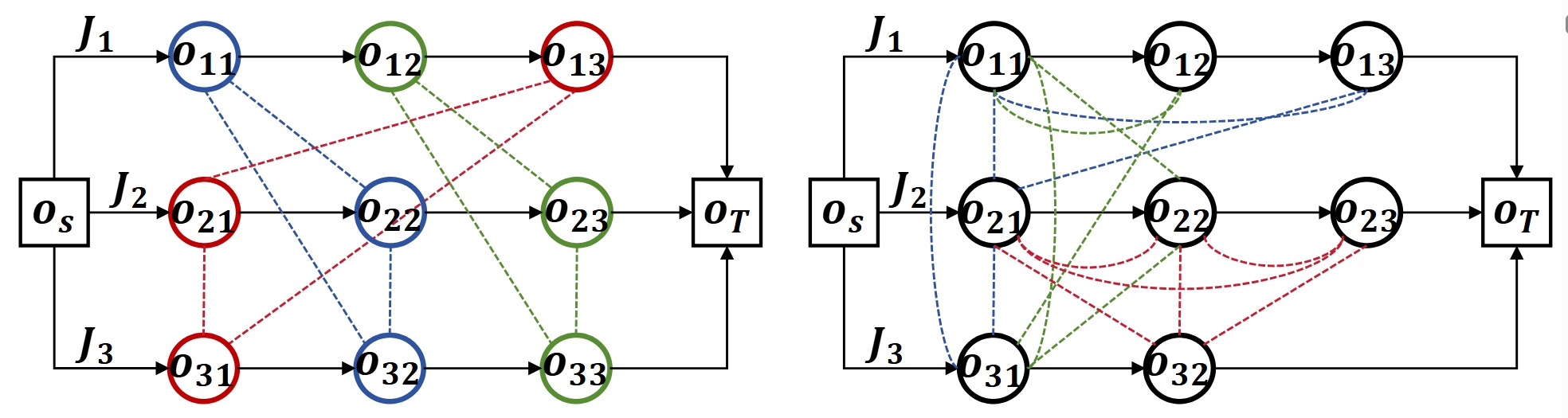}
\caption{Disjunctive graph representations of JSP and FJSP with 3 jobs and 3 machines. The arrows on black lines indicate precedence within jobs, while dotted lines represent disjunctive arcs, whose directions must be determined to establish a valid schedule. Disjunctive arcs of the same color indicate the linked operations should be executed by the same machine(s).}
\label{fig:illustration}
\end{figure}
The job shop scheduling problem (JSSP) aims to schedule \( n \) jobs \( J = \{J_1, J_2, \ldots, J_n\} \) across \( m \) machines \( M = \{M_1, M_2, \ldots, M_m\} \). Each job \( J_i\in J \) comprises a series of operations, where $O_{ij}$ represents the $j$-th operation in the $i$-th job. In the standard JSSP, each operation $O_{ij}$ must be processed on a specific machine $M_k$ with a processing time $p_{ij}^{k}$. Operations should be scheduled so that each machine handles only one operation at a time, and the operations within each job are executed in their designated sequence. The objective is to optimally assign operations to machines to minimize the makespan, i.e., the overall processing time for all operations.

In the flexible job shop problem (FJSP), each operation \( O_{ij} \) can be processed on one or more machines. Compared to JSSP, this flexibility introduces an additional layer of optimization complexity, and an effective solution should strategically utilize machines to minimize the makespan. The disjunctive graph representations of JSSP and FJSP are shown in Figure~\ref{fig:illustration}.


\begin{figure*}[!ht]
\centering
\includegraphics[width=0.95\textwidth]{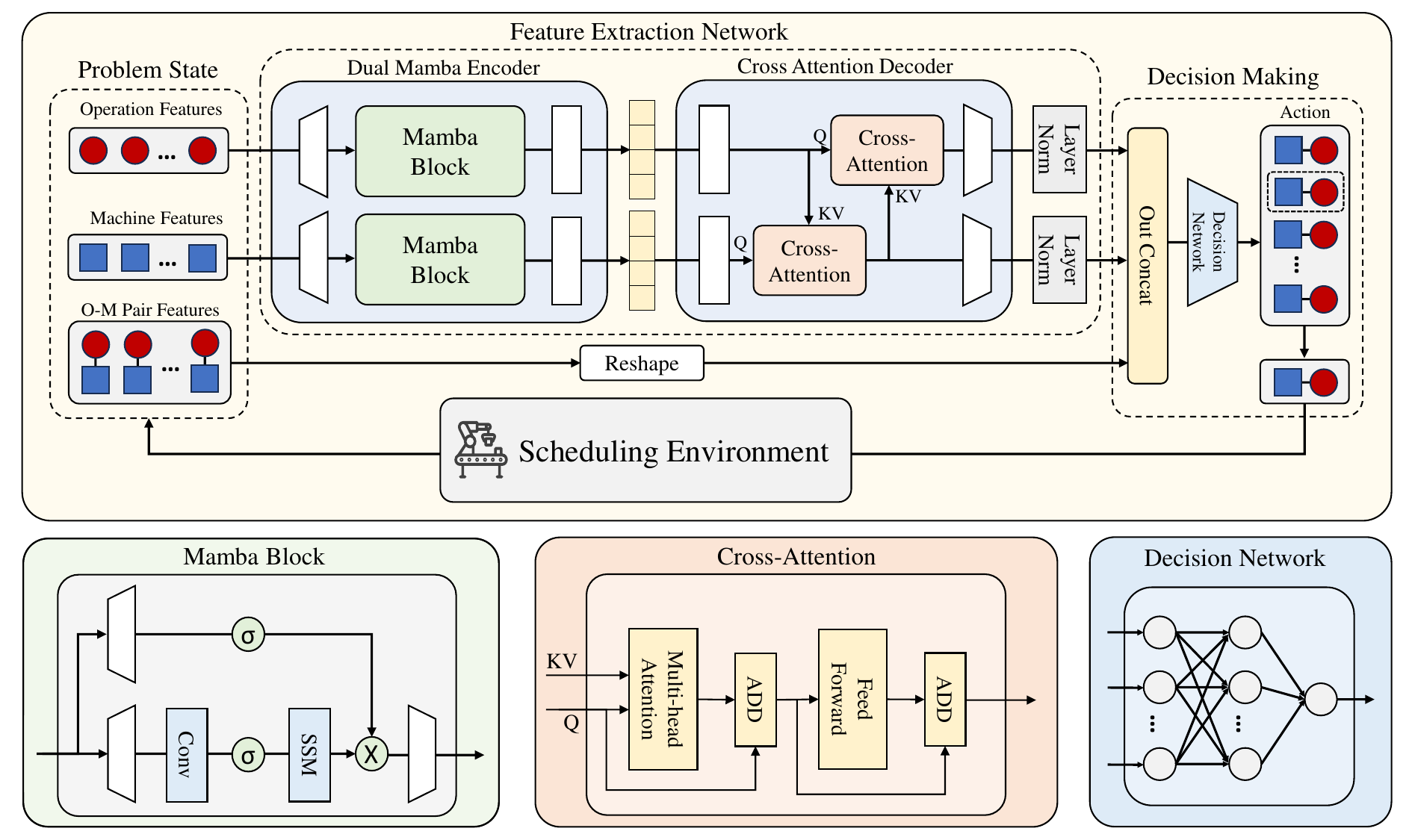}
\caption{
The overall architecture of our proposed Mamba-CrossAttention framework. It comprises two fundamental components: the feature extraction network and the decision-making network. At each step in the scheduling process, the feature extraction network extracts raw features of operations and machines from the environment. The decision network concatenates the output of the feature extraction network with the machine-operation pair features to form candidates, and selects the optimal O-M pair according to the probability to complete the end-to-end solution generation.
}
\label{Mamba-CrossAttention}
\end{figure*}

\section{Methodology}
Current research predominantly relies on either graph convolution or graph attention mechanisms as the fundamental approach for learning the embeddings of machines and operations for FJSP,
which typically necessitates a specialized graph-based state configuration of the problem. Nonetheless, the embeddings learned from local regions surrounding the target (operation or machine) often require K-hop message passing, typically instantiated as a neural network with K layers where each layer represents the graph convolution or graph attention operator, to acquire the global context of the graph. This framework is susceptible to over-smoothing challenges \cite{rusch2023survey} and struggles to capture the holistic operation and machine sequences for FJSP. In this paper, we show that learning the operation and machine embeddings from the whole sequence is more effective (i.e., yielding better performance) and more efficient with the linear state-space model. We start by first modelling the FJSP problem as a sequential decision-making problem.

\subsection{MDP Formulation}  
The process of constructing a solution to an FJSP instance can be viewed as a sequential decision-making process where at each decision step $t$ (and $T(t)$ is the wall-clock time), we dispatch an available operation-machine pair $(O_{ij}, M_k)$, indicating that the operation $O_{ij}$ is selected to be scheduled and processed on its compatible machine $M_k$ at $t$. This process continues until all operations have been scheduled. The formal Markov Decision Process (MDP) for FJSP is defined as follows.

\noindent\textbf{\underline{Action}.}
The action space $\mathcal{A}(t)$ at step $t$ is defined as the set of operations ready to be scheduled and their corresponding eligible machines. They are arranged in operation-machine pairs.
A pair $(O_t, M_t) \in \mathcal{A}(t)$ if and only if the operation $O_t$ is available for processing and there is an eligible machine $M_k$ that can execute $O$ at $t$. Specifically, an operation $O_{ij}$ from job $J_i$ is considered eligible for processing at $t$ if its immediate predecessor, $O_{ij-1}$, has already been completed, and there exists an idle machine $M_k$ ready for processing $O_{ij}$. As the scheduling process progresses, the number of unscheduled operations gradually decreases, continuing until all operations have been scheduled. Therefore, the decision action is performed $|\mathcal{O}|$ times, with $|\mathcal{O}|$ denoting the total number of operations, as each operation requires to be scheduled only once.

\noindent\textbf{\underline{State}.}
The state $s_t$ encompasses the overall status and configurations of operations and machines at $t$ (e.g., processing duration of operation and machine available time). We use the handcraft features to represent $s_t$. Specifically, let
$\mathcal{O}(t)$, $\mathcal{M}(t)$, and $\mathcal{A}(t)$ be the set of all operations, all machines, and the actions at $t$.
Then, $s_t$ consists of features from three parties, including $h_{O_{ij}} \in \mathbb{R}^{o}$ for each operation $O_{ij}\in \mathcal{O}(t)$, $h_{M_k} \in \mathbb{R}^{m}$ for each machine $M_k \in \mathcal{M}(t)$, and $h_{(O_{ij},M_k)} \in \mathbb{R}^{a}$ for each eligible operation-machine pair $(O_{ij},M_k) \in \mathcal{A}(t)$.
For operations already processed at $t$, we employ a dummy zero feature vector, as they will not affect the current scheduling. 

\noindent\textbf{\underline{State transition}.}
After invoking the action $a_t$ in $s_t$, the configurations of the scheduling environment are changed (e.g., new machines become available). The operation features $h_{O_{ij}}$, machine features $h_{M_k}$, and action features $h_{(O_{ij}, M_k)}$ are updated accordingly, resulting in the next state $s_{t+1}$. Due to limited space, we refer the audiences to Appendix \ref{app:raw-feature} for the details of the state vectors and transitions.

\noindent\textbf{\underline{Reward}.}
The difference in makespan between states $s_t$ and $s_{t+1}$ can be used to estimate the quality of the decision $a_t$ made at $t$. Thus, we define the reward for $(s_t, a_t)$ as $r(s_t,a_t,s_{t+1})=C_{\max}(s_t)-C_{\max}(s_{t+1})$. It is easy to show that the cumulative reward $G=\sum_{t=0}^{|\mathcal{O}|}r(S_t,a_t, S_{t+1})=C_{\max}(s_0)-C_{\max}$. Since the makespan $C_{\max}(s_0)$ of the initial state $s_0$ is constant, maximizing the cumulative reward $G$ can be equivalent to minimizing the total completion time $C_{\max}$.

\noindent\textbf{\underline{Policy}.}
The M-CA policy network $\pi_\theta(a_t|s_t)$, parameterized by trainable parameters $\theta$, generates the probability of selecting action $a_t$ at state $s_t$. It consists of a feature extraction network and a decision-making network. To find an optimal $\theta$, we developed an actor-critic algorithm. Specifically, the actor selects actions based on $\pi_\theta(a_t|s_t)$, while the critic estimates the cumulative reward $G$ and provides feedback to update the policy. The next section details $\pi_\theta(a_t|s_t)$ and the actor-critic algorithm for optimizing $\theta$.

\subsection{Mamba-CrossAttention Policy Network}
The architecture of the Mamba-CrossAttention (M-CA) model is illustrated in Figure~\ref{Mamba-CrossAttention}. It comprises two fundamental components: the feature extraction network and the decision-making network. The feature extraction network follows an encoder-decoder structure. The encoder integrates dual Mamba blocks to extract operation and machine embeddings independently. These embeddings are subsequently fed into the cross-attention decoder to acquire interactive embeddings of machines and operations. Subsequently, the interactive features and the reshaped operation-machine pair features are concatenated and directed to the decision-making network to determine the probability associated with selecting each eligible operation-machine pair. As per our action selection strategy, the pair with the highest probability is chosen under the greedy strategy, while under the sampling strategy, selection is proportional to the probability distribution.

\subsubsection{Feature Extraction Network} \hfill\\
\noindent\textbf{Encoder.} 
The Dual Mamba Encoder (DME) is designed to extract the raw features of machines and operations separately. 
For all operations raw feature sequence $H_{O_{u}}=\{\ldots,h_{O_{ij}},\ldots\}$ and machine raw feature sequence $H_{M_{u}}=\{\ldots,h_{M_{k}},\ldots\}$, we use linear projections to expand their dimensions and feed them into Mamba blocks. Mamba follows the residual structure. In the Mamba block, the main branch uses a convolution layer to extract local features, which are then activated and fed into the SSM. The residual branch of inputs are processed using an activation function (SiLU \cite{hendrycks2016gaussian}). After that, the main branch and the res  
 idual branch are connected in a nonlinear manner (i.e., multiplication) to obtain the processed features. The computational process of the DME is as follows:
\begin{equation}
\label{eq:mamba}
    \begin{aligned}
    &H^{\prime}_{O_{u}} = \text{SSM}(\sigma(\text{Conv}(\text{Linear}(H_{O_{u}}))) + \sigma(\text{Linear}(H_{O_{u}})),\\
    &H^{\prime}_{M_{u}} = \text{SSM}(\sigma(\text{Conv}(\text{Linear}(H_{M_{u}}))) + \sigma(\text{Linear}(H_{M_{u}})).
    \end{aligned}
\end{equation}
Mamba focuses on specific machines or operations in the full sequence rather than on all features during feature extraction by integrating the selective scanning mechanism in SSM. In other words, Mamba achieves a balance between feature extraction capability and computational efficiency through selectivity. In contrast to graph attention mechanisms, Mamba does not need to model the neighborhood relationships between operations or machines manually but instead focuses on the implicit connections between them. With the SSM inherence and the selective scan mechanism, it achieves linear time complexity $O(|\mathcal{O}|)$ and $O(|\mathcal{M}|)$ in processing the operation sequence $H_{O_{u}}$ and machine sequence $H_{M_{u}}$. Therefore, Mamba is a simple and effective way to extract features for scheduling problems. In addition, the Dual Mamba Encoder can be used alone for feature extraction, in which case the output features of Equation \ref{eq:mamba} are used directly as inputs to the decision module.

\noindent\textbf{Decoder.} The Cross-Attention Decoder is designed to simulate two-way selection between machines and operations during the scheduling process. The process of $\text{CrossAttention}$ is as follows, which includes a multi-head attention layer and a feed-forward layer:
\begin{equation}
\begin{aligned}
&\hat{\mathbf{h}}_i=\mathbf{h}_i+\mathrm{MHA}(\mathbf{h_i},E),
\\&\mathbf{h}_i^{\prime}=\hat{\mathbf{h}}_i+\mathrm{FF}(\hat{\mathbf{h}}_i).
\end{aligned}
\end{equation}
where $H=(\mathbf{h_1},\ldots,\mathbf{h_n})$, $E=(\mathbf{e_1},\ldots,\mathbf{e_n})$ denotes the input feature matrices. $H$ denotes the $Q$ matrix and $E$ denotes the $K$ and $V$ matrix.

We utilize the cross-attention mechanism in an efficient way. Our decoder consists of two cross-attention layers. After encoding, we obtain the operation embedding matrix $H^{\prime}_{O_{u}}$ and the machine embedding matrix $H^{\prime}_{M_{u}}$. We use $H^{\prime}_{M_{u}}$ as the input to the $Q$ matrix of the first cross-attention layer and $H^{\prime}_{O_{u}}$ as the $K$ and $V$ matrices to obtain the machine matrix $H^{\prime\prime}_{M_{u}}$ containing the features of the operation. This process simulates the selection of machines for operations. After that, we input the matrix $H^{\prime}_{O_{u}}$ as $Q$ and the matrix $H^{\prime\prime}_{M_{u}}$ as $K$ and $V$ into the second cross-attention layer to obtain the operation matrix $H^{\prime\prime}_{O_{u}}$, which contains the machine features. This process simulates the selection of operations for machines. And we use LayerNorm to stabilize training. The whole process in the decoder can be represented as follows:

\begin{equation}
\begin{aligned}
H^{\prime\prime}_{M_{u}}=\text{CrossAttention}(H^{\prime}_{M_{u}},H^{\prime}_{O_{u}}),\\
H^{\prime\prime}_{O_{u}}=\text{CrossAttention}(H^{\prime}_{O_{u}},H^{\prime\prime}_{M_{u}}).
\end{aligned}
\end{equation}
Through the above process, we achieve the implicit interaction between machine and operation in the state feature extraction process, and complete the fusion of machine and operation features. Although the operation-machine allocation is made by the decision network, we argue that their hidden connections need to be considered at a global level during the feature extraction phase. Therefore, we propose the cross-attention decoder. Additionally, the traditional self-attention mechanism has $O(|\mathcal{O}|^2)$ computational complexity when extracting operation features, i.e., each element in the sequence has to pay attention to all the elements in the sequence, which does not allow for interaction between different feature sequences. In contrast, our cross-attention decoder has a lightweight complexity of $O(|\mathcal{O}|\times|\mathcal{M}|)$, and the operation feature $H^{\prime}_{O_{u}}$ can interact with both operation and machine features in $H^{\prime\prime}_{M_{u}}$ at the second cross-attention layer, which combines both self-attention and cross-attention features. Considering that the number of operations $|\mathcal{O}|$ is much larger than the number of machines $|\mathcal{M}|$ in scheduling problems in manufacturing (i.e. more jobs and a fixed number of machines),  the complexity of our method tends to be linearly related to $|\mathcal{O}|$. Combining the linear-time encoder and the lightweight decoder, our approach has the potential to be applied to solve large-scale problems in real-world production scenarios.

\subsubsection{Decision-Making Network}\hfill\\
After processing the raw features of the operations and machines with the feature extraction network, we use non-zero averaging for pooling to obtain the global features $h_{G(O)}^{\prime\prime}$ and $h_{G(M)}^{\prime\prime}$ of the machines and operations. The averaging process is as follows:
\begin{equation}
\begin{aligned}
&h_{G(O)}^{\prime\prime}=\frac{1}{\left| {O_{ij}\in \mathcal{O}_{u}}| h_{O_{ij}}^{\prime\prime} \neq 0\right | }\sum_{O_{ij}\in \mathcal{O}_{u},h_{O_{ij}}^{\prime\prime} \neq 0} h_{O_{ij}}^{\prime\prime},  h_{O_{ij}}^{\prime\prime} \in H^{\prime\prime}_{O_{u}},\\
&h_{G(M)}^{\prime\prime}= \frac{1}{\left | M_k\in \mathcal{M}_{u} | h_{M_{k}}^{\prime\prime} \neq 0 \right | }\sum_{M_{k}\in \mathcal{M}_{u},h_{M_{k}}^{\prime\prime} \neq 0} h_{M_{k}}^{\prime\prime}, h_{M_{k}}^{\prime\prime} \in H^{\prime\prime}_{M_{u}}.
 \end{aligned}
\end{equation}

Then, we concatenate the operation features $h_{O_{ij}}^{\prime\prime}$,machine features $h_{M_k}^{\prime\prime}$, global features $h_{G(O)}^{\prime\prime}$ and $h_{G(M)}^{\prime\prime}$ with the operation-machine pair features $h_{(O_{ij},M_k)}$ to form the complete decision network's candidate feature $h_c{(O_{ij},M_k)}$.
\begin{equation}
	h_c{(O_{ij},M_k)}=h_{O_{ij}}^{\prime\prime} \big\|h_{M_k}^{\prime\prime} \big\| h_{G(O)}^{\prime\prime}  \big\| h_{G(M)}^{\prime\prime} \big\|h_{(O_{ij},M_k)}.
\end{equation}
Finally, the probability of selecting each candidate (i.e., O-M pair) is obtained by the actor network $\text{MLP}_\theta$.
\begin{equation}
	\pi_\theta\left(a_t \mid s_t\right)=\text{Softmax}(\operatorname{MLP}_\theta(h_c{(O_{ij},M_k)})).
\end{equation}
where $\pi_\theta\left(a_t \mid s_t\right)$ denotes the probability distribution of choosing action $a_t$ in state $s_t$.



\subsection{Training Algorithm}
In this research, we utilize the proximal policy optimization (PPO) algorithm \cite{schulman2017proximal} to effectively train the proposed scheduling model.
It follows the actor-critic architecture, containing the actor network $MLP_\theta$ and the critic network $MLP_\phi$. $MLP_\theta$ is used to generate action probabilities during training and testing, while $MLP_\phi$ is only used to evaluate the model during training.
To ensure training stability, we employ the generalized advantage estimation (GAE) technique \cite{schulman2015high}. 
The model is trained for $N$ total iterations, with every iteration containing a batch of $B$ instances. In each iteration, the DME or M-CA policy interacts with a batch of FJSP environments $E$ of the same scale instances in parallel, gathering transition data for updating the model parameters $\Theta$ later. The environments are refreshed every $N_{res}$ batches based on a fixed distribution. Additionally, the policy undergoes validating on predetermined validation environments $E_v$ for every $N_{val}$ batch. The validation datasets are generated under the same distribution as the training data. In our experiments, we combine two distinct strategies for action selection. The first strategy involves a greedy approach, where the action with the highest probability is consistently chosen, primarily utilized during validation for high-confident evaluation results. The second strategy entails an action-sampling approach, where actions are sampled from the distribution $\pi_\theta$ during training to ensure adequate exploration. The detailed algorithm is given in Appendix~\ref{app:Algorithm}.

\begin{table*}[!ht]
\setlength{\tabcolsep}{4pt}
\caption{Results on benchmarks}
\label{tab:results3}
\fontsize{9}{10}\selectfont
\begin{tabular}{l|cc|cc|cc|cc|cc|cc}
\toprule
         & \multicolumn{2}{c|}{Brandimarte}         & \multicolumn{2}{c|}{Hurink(rdata)}           & \multicolumn{2}{c|}{Hurink(edata)}           & \multicolumn{2}{c|}{Hurink(vdata)}      & \multicolumn{2}{c|}{Barnes}                  & \multicolumn{2}{c}{Dauzere}                 \\
         & Obj/Gap          & Time(s)                 & Obj/Gap              & Time(s)                 & Obj/Gap              & Time(s)                 & Obj/Gap         & Time(s)                 & Obj/Gap              & Time(s)                 & Obj/Gap              & Time(s)                 \\
\midrule
OR-Tools & 174.20           & 1447                 & 935.80               & 1397                 & 1028.93              & 900                  & 919.60          & 639                  & 995.19               & 879                  & 2224.17              & 1701                 \\
         & 1.50\%           &  & 0.11\%               &  & -0.03\%              &  & -0.01\%         &  & -0.18\%              &  & 0.52\%               &  \\
\midrule
FIFO     & 205.56           & 0.16                 & 1087.12              & 0.17                 & 1244.92              & 0.17                 & 982.89          & 0.17                 & 1281.57              & 0.19                 & 2564.67              & 0.35                 \\
         & 31.82\%          &  & 17.25\%              &  & 20.83\%              &  & 7.58\%          &  & 28.77\%              &  & 15.63\%              &  \\
MOR      & 200.36           & 0.16                 & 1066.73              & 0.17                 & 1227.07              & 0.17                 & 966.01          & 0.17                 & 1231.52              & 0.19                 & 2478.56              & 0.38                 \\
         & 28.08\%          &  & 15.07\%              &  & 19.24\%              &  & 5.68\%          &  & 23.89\%              &  & 11.61\%              &  \\
SPT      & 237.52           & 0.16                 & 1200.41              & 0.16                 & 1312.84              & 0.16                 & 1082.88         & 0.17                 & 1245.67              & 0.19                 & 2765.61              & 0.35                 \\
         & 44.88\%          &  & 29.47\%              &  & 26.79\%              &  & 18.2\%          &  & 24.43\%              &  & 24.99\%              &  \\
MWKR     & 201.74           & 0.16                 & 1053.10              & 0.17                 & 1219.01              & 0.17                 & 952.01          & 0.17                 & 1200.19              & 0.19                 & 2455.61              & 0.35                 \\
         & 28.91\%          &  & 13.86\%              &  & 18.6\%               &  & 4.22\%          &  & 20.59\%              &  & 10.58\%              &  \\

\midrule
MLP*     & 186.80           & 0.41                 & 1041.83              & 0.28                 & 1187.93              & 0.27                 & 963.50          & 0.27                 & \textbf{1133.62}     & 0.39                 & 2463.22              & 0.41                 \\
         & 14.04\%          &  & 12.14\%              &  & 15.54\%              &  & 5.35\%          &  & \textbf{13.83\%}     &  & 11.02\%              &  \\
HGNN-G    & 199.80           & 0.86                 & 1039.65              & 0.94                 & 1205.98              & 0.94                 & 956.43          & 0.95                 & 1183.14              & 1.02                 & 2436.00              & 1.94                 \\
         & 25.43\%          &  & 11.78\%              &  & 17.11\%              &  & 4.72\%          &  & 18.57\%              &  & 9.74\%               &  \\
HGNN-S    & 192.60           & 2.53                 & 998.98               & 2.86                 & 1129.28              & 2.89                 & 932.10          & 2.78                 & 1108.71              & 3.06                 & 2373.61              & 7.53                 \\
         & 19.97\%          &  & 7.04\%               &  & 9.42\%               &  & 1.47\%          &  & 11.22\%              &  & 7.07\%               &  \\
DAN-G    & 188.60           & 0.93                 & 1031.68              & 0.98                 & \textbf{1178.93}     & 0.97                 & 942.63          & 1.27                 & 1181.71              & 1.04                 & 2394.06              & 1.95                 \\
         & 16.5\%           &  & 11.13\%              &  & \textbf{14.66\%}     &  & 2.9\%           &  & 18.58\%              &  & 7.92\%               &  \\
DAN-S    & 182.20           & 3.07                 & 981.93               & 3.99                 & 1122.73              & 3.50                 & 925.00          & 4.35                 & 1109.00              & 4.34                 & 2338.28              & 8.98                 \\
         & 9.93\%           &  & 5.33\%               &  & 9.1\%                &  & 0.68\%          &  & 11.27\%              &  & 5.45\%               &  \\
\midrule
DME-G  & 181.60           & 0.44                 & 1030.00              & 0.45                 & 1204.80              & 0.46                 & \textbf{941.88} & 0.46                 & 1169.10              & 0.53                 & 2396.83              & 0.96                 \\
    & 10.88\%          &  & 11.16\%              &  & 17.1\%               &  & \textbf{2.83\%} &  & 17.15\%              &  & 8.01\%               &  \\
DME-S  & 178.90           & 2.43                 & 982.60               & 2.85                 & 1133.53              & 2.97                 & 925.55          & 2.99                 & 1117.71              & 3.63                 & 2347.06              & 7.77                 \\
     & 7.24\%           &  & 5.47\%               &  & 10.33\%              &  & 0.77\%          &  & 12.15\%              &  & 5.82\%               &  \\
M-CA-G   & \textbf{181.00}  & 0.57        & \textbf{1020.88}     & 0.58                 & 1183.78              & 0.60                 & 943.23          & 0.60                 & 1143.00              & 0.65                 & \textbf{2382.72}     & 1.20                 \\
    & \textbf{10.23\%} &  & \textbf{9.8\%}       &  & 15.55\%              &  & 2.98\%          &  & 14.78\%              &  & \textbf{7.38\%}      &  \\
M-CA-S   & \textbf{177.90}  & 2.66                 & \textbf{975.90}      & 3.12                 & \textbf{1119.03}     & 3.26                 & \textbf{924.88} & 3.27                 & \textbf{1103.05}     & 4.00                 & \textbf{2329.67}     & 8.43                 \\
    & \textbf{6.78\%}  &  & \textbf{4.61\%}      &  & \textbf{8.65\%}      &  & \textbf{0.65\%} &  & \textbf{10.73\%}     &  & \textbf{5.05\%}      & \\
\bottomrule
\end{tabular}

\begin{tablenotes} \small
\item[\textbf{2}] The best results of greedy and sampling are highlighted by bold values.
 	 	\item[\textbf{1}] The results of the methods marked with (*) are obtained directly from the original paper. 
  \end{tablenotes}
  
\end{table*}

\section{Experiments}
\subsection{Benchmarks}
In this work, we evaluate our approaches with six public benchmarks widely used in previous work \cite{song2023flexible,wang2024flexible,yuan2024solving}. They are Barnes dataset \cite{barnes1996flexible}, Brandimarte dataset \cite{brandimarte1993routing}, Dauzere dataset \cite{dauzere1997integrated}, and 3 groups of Hurink dataset \cite{hurink1994tabu}. We give more details of the benchmark data in Appendix \ref{app:benchmarks-detail}. As for the synthetic data, we used the instances from \cite{song2023flexible} for testing, spanning six problem sizes, $10\times5$, $20\times5$, $15\times10$, $20\times10$, $30\times10$ and $40\times10$, each of which consists of 100 instances. In these data, the processing time $p_{ij}^{k}$ for each operation $O_{ij}$ on machine $M_k$ is uniformly sampled from $U(1,20)$. 
We also test the generalization performance by evaluating our method on the extremely large instances that are randomly generated with sizes from $100\times10$ to $10000\times10$. More details are provided in Appendix \ref{app:super-size}.

\begin{figure}[h]
\centering
\includegraphics[width=0.47\textwidth]{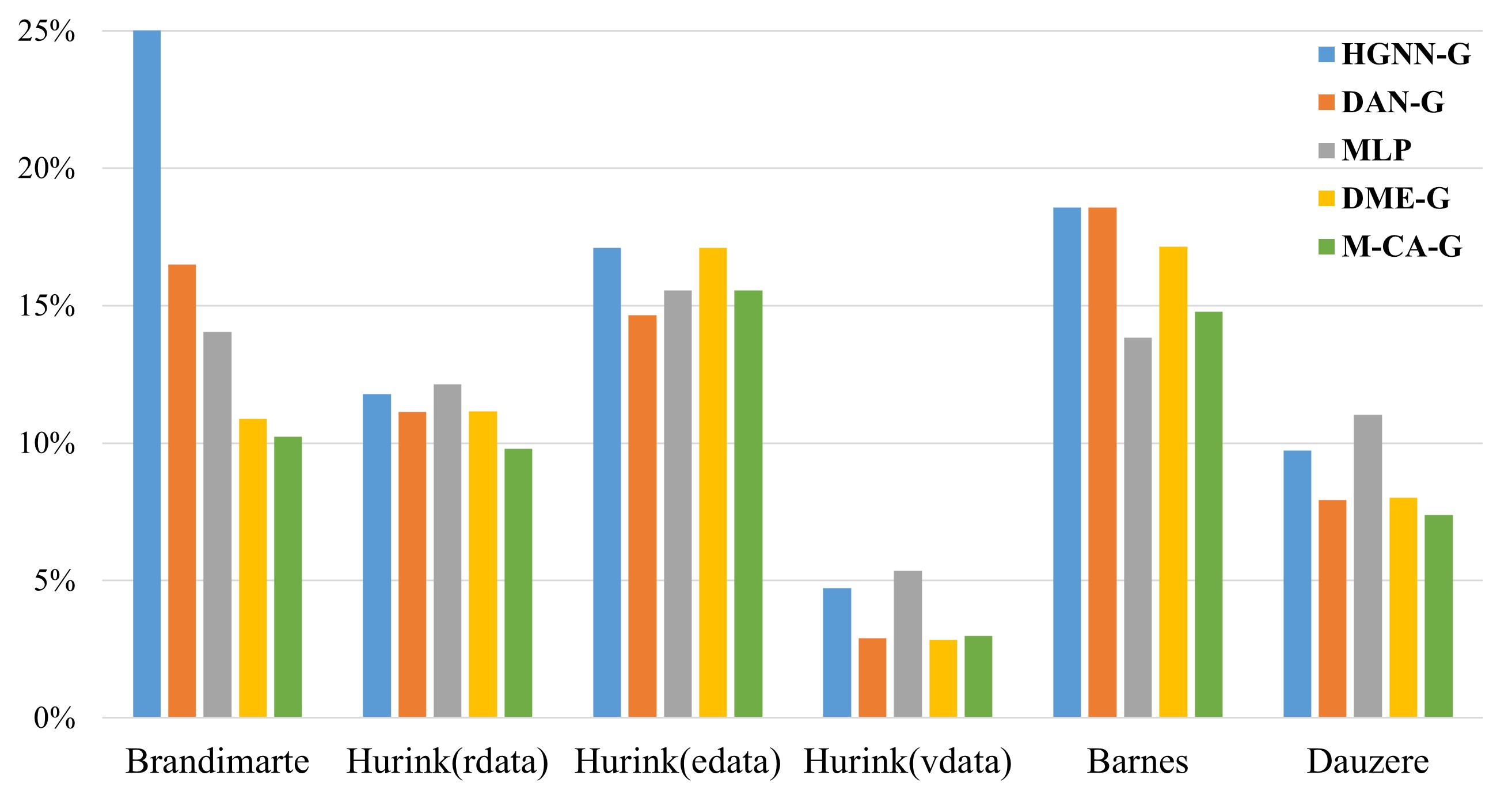}
\caption{Average gaps on Benchmarks with greedy strategy.}
\label{Benchmarks}
\end{figure}

\subsection{Baselines}
We conducted extensive experiments to thoroughly evaluate the performance of our method by comparing it against a series of state-of-the-art baselines, including
(1) Exact Solver: Google OR-Tools \cite{ortools}, where we set an 1800-second time limit across all benchmarks. However, for the extremely large instances in the generalization experiment, we set a longer time limit for OR-Tools for better solutions; (2) The priority dispatching rules (PDRs) that are widely used in real industries: FIFO (first in first out), MOPNR (most operations remaining), SPT (shortest processing time) and MWKR (most work remaining). We refer the audiences to Appendix \ref{app:pdr} for the detailed description and definition of these PDRs; (3) Learning-based methods (categorized by representation learning approach): Heterogeneous graph based HGNN \cite{song2023flexible}, Graph attention based DAN \cite{wang2024flexible} and lightweight MLP\cite{yuan2024solving}.

\begin{table*}[t]
\setlength{\tabcolsep}{4pt}
\centering
\caption{Results on synthetic data}
\label{tab:results1}
\fontsize{9}{9}\selectfont
\begin{tabular}{l|ccc|ccc|ccc|ccc}
\toprule
           & \multicolumn{3}{c|}{FJSP 10x5}               & \multicolumn{3}{c|}{FJSP 20x5}               & \multicolumn{3}{c|}{FJSP 15x10}              & \multicolumn{3}{c}{FJSP 20x10}             \\
           & Obj             & Gap             & Time(s)    & Obj             & Gap             & Time(s)    & Obj             & Gap             & Time(s)    & Obj             & Gap              & Time(s)    \\
\midrule
OR-Tools   & 96.32           & 0.00\%          & 1745 & 188.15          & 0.00\%          & 1800 & 143.53          & 0.00\%          & 1696 & 195.98          & 0.00\%           & 1805 \\
\midrule
FIFO       & 119.4           & 24.06\%         & 0.07    & 216.08         & 14.87\%         & 0.10    & 184.55         & 28.65\%         & 0.16    & 233.48         & 19.22\%          & 0.23    \\
MOR        & 115.38         & 19.87\%         & 0.05    & 214.16          & 13.85\%         & 0.10    & 173.15         & 20.68\%         & 0.17    & 219.8         & 12.2\%           & 0.23    \\
SPT        & 129.82         & 34.76\%         & 0.05    & 230.48         & 22.56\%         & 0.10    & 198.33         & 38.22\%         & 0.17    & 255.17          & 30.25\%          & 0.23    \\
MWKR       & 113.23         & 17.58\%         & 0.05    & 209.78         & 11.51\%         & 0.10    & 171.25          & 19.41\%         & 0.18    & 216.11         & 10.3\%           & 0.23    \\
\midrule
HGNN-G      & 112.47          & 16.82\%         & 0.30    & 211.77          & 12.57\%         & 0.58    & 165.39          & 15.29\%         & 0.94    & 215.93          & 10.22\%          & 1.25    \\
HGNN-S      & 106.19          & 10.30\%         & 0.74    & 210.5           & 11.88\%         & 1.47    & 160.34          & 11.77\%         & 2.42    & 215.88          & 10.18\%          & 3.55    \\
DAN-G      & 108.76          & 12.96\%         & 0.32    & 197.03          & 4.72\%          & 0.63    & 160.61          & 11.9\%          & 0.97    & 199.73          & 1.95\%           & 1.45    \\
DAN-S      & 102.21          & 6.13\%          & 0.52    & 193.89          & 3.06\%          & 1.27    & 152.34          & 6.17\%          & 2.82    & 195.14          & -0.4\%           & 4.97    \\
\midrule
DME-G    & 106.17          & 10.24\%         & 0.15    & 193.36          & 2.8\%           & 0.30    & 157.91          & 10.05\%         & 0.46    & \textbf{193.65} & \textbf{-1.17\%} & 0.61   \\
DME-S    & 100.79          & 4.67\%          & 0.34    & 190.46          & 1.24\%          & 0.96    & 149.93          & 4.49\%          & 2.26    & 190.13          & -2.96\%          & 4.26    \\
M-CA-G & \textbf{105.8}  & \textbf{9.85\%} & 0.19    & \textbf{193.12} & \textbf{2.66\%} & 0.38    & \textbf{157.56} & \textbf{9.78\%} & 0.58    & 193.9           & -1.03\%          & 0.78    \\
M-CA-S & \textbf{100.41} & \textbf{4.24\%} & 0.38    & \textbf{190.21} & \textbf{1.1\%}  & 1.05    & \textbf{149.85} & \textbf{4.44\%} & 2.47    & \textbf{189.83} & \textbf{-3.11\%} & 4.61  \\
\bottomrule
\end{tabular}
\end{table*}
\begin{table*}[h]
\setlength{\tabcolsep}{2pt}
\centering
\caption{Generalization on large size instances}
\label{tab:results2}
\fontsize{9}{10}\selectfont
\begin{tabular}{l|ccc|ccc|ccc|ccc|ccc}
\toprule
\multicolumn{1}{c|}{} & \multicolumn{3}{c|}{FJSP 30x10}        & \multicolumn{3}{c|}{FJSP 40x10}           & \multicolumn{3}{c|}{FJSP 100x10}          & \multicolumn{3}{c|}{FJSP 1000x10}             & \multicolumn{3}{c}{FJSP 10000x10} \\
\multicolumn{1}{c|}{} & Obj          & Gap             & Time(s) & Obj             & Gap             & Time(s) & Obj            & Gap              & Time(s) & Obj             & Gap               & Time(s)   & Obj              & Gap   & Time   \\
\midrule
OR-Tools             & 274.67       & 0.00\%          & 1772 & 365.96          & 0.00\%          & 1769 & 940.7          & 0.00\%           & 1800 & 10283.9         & 0.00\%            & 3603   & -                & OOM-CPU     & -      \\
\midrule
FIFO                 & 328.238      & 19.56\%         & 0.37 & 426.968         & 16.7\%          & 0.51 & 1045.7         & 11.17\%          & 1.72 & 10302           & 0.24\%            & 87.28  & 102534.6           & -     & 2.72h  \\
MOR                  & 317.428      & 15.6\%          & 0.37 & 421.344         & 15.16\%         & 0.52 & 1043           & 10.88\%          & 1.73 & 10296.8         & 0.18\%            & 86.13  & 102542.8           & -     & 2.85h  \\
SPT                  & 350.074      & 27.47\%         & 0.36 & 445.17          & 21.66\%         & 0.51 & 1026           & 9.06\%           & 1.71 & 9570.7          & -6.88\%           & 84.22  & 94728.4            & -     & 2.71h  \\
MWKR                 & 312.926      & 13.96\%         & 0.37 & 414.816         & 13.37\%         & 0.52 & 1038.7         & 10.42\%          & 1.74 & 10273           & -0.04\%           & 85.12  & 102508           & -     & 2.76h  \\
\midrule
HGNN                  & 313.79       & 14.27\%         & 1.82 & 416.56          & 13.85\%         & 2.57 & 1043.2         & 10.90\%          & 6.26 & 10287.1         & -33.01\%          & 1020   &         -       & OOM-GPU   &   -   \\
DAN                  & 283.57       & 3.27\%          & 2.24 & 373.22          & 2.0\%           & 2.80 & 924.8          & -1.68\%          & 8.69 & 9157.8          & -40.36\%          & 140.30 & 93871.8            & -     & 3.46h  \\
\midrule
DME                & 277.17       & 0.94\%          & 0.93 & \textbf{366.44} & \textbf{0.15\%} & 1.28 & \textbf{913.9} & \textbf{-2.84\%} & 3.51 & 9617.5          & -37.38\%          & 106.10 & 99804.4           & -     & 3.10h  \\
M-CA                 & \textbf{277} & \textbf{0.87\%} & 1.17 & 367.06          & 0.32\%          & 1.59 & 916.8          & -2.54\%          & 4.52 & \textbf{9038.1} & \textbf{-41.15\%} & 120.44 & \textbf{90200}   & -     & 3.34h\\
\bottomrule
\end{tabular}
\begin{tablenotes} \small
 	 	\item[\textbf{2}]  "OOM" indicates that the method is out of memory. 
        \item[\textbf{2}] "GPU" indicates the video memory (24GB), "CPU" indicates the system memory (128GB).    
  \end{tablenotes}
\end{table*}
\subsection{Experiment Configurations}

We used the same hyperparameter settings for all problem sizes. We train the model for 10,000 iterations, each iteration containing 20 independent trajectories (i.e., instances). The instances used for training are regenerated every 20 iterations. The model is validated on 100 instances, which are generated using the same rules as the training instances and fixed during training. We validate the model performance every 10 iterations and store the best weights. The learning rate was set to $1\times10^{-4}$ and decayed to $1\times10^{-5}$ halfway through the training process. 
The training settings are the same when training the DME and MCA. For the Mamba structure, we use its classic Mamba \cite{gu2023mamba} setting. We simply set the number of Mamba blocks used for operation and machine features extraction to 1 and empirically observed strong performance. The model dimension (i.e., the embedding dimension) is set to 128. For cross-attention, we set the head number of multi-head attention to 8. For the decision network, both actor and critic networks are 3-layer MLPs, with hidden layers of dimension 64. Our training and testing experiments were conducted with a single RTX 3090 GPU. For the comparison methods, we followed their original settings to reproduce them. 

\subsection{Results}
\noindent\textbf{Results on benchmarks.} First, to explore the cross-distribution generalizability of our approaches, we test them on various public benchmarks. We directly used the model trained on the problem size of $10\times5$ to test and observe promising results. For HGNN and DAN, we also used models trained on the same size, and some of the results were better than their original papers. For MLP, due to the lack of code, we directly used the results from their paper, which was also trained on size $10\times5$. 
As popular, for each problem size we report the average objective value (i.e., makespan), the average optimal solution gap, and the average time (second) to obtain a solution. The optimal solution gap $g$ is used to measure the gap between the solution's makespan $C$ and the best-known solution's makespan $C_{BS}$, which is calculated as follows:
\begin{equation}g=
\begin{pmatrix}
C/C_{BS}-1
\end{pmatrix}\times100\%.
\end{equation}
For public benchmarks, we use the best-known upper bounds from their original paper as best-known solutions. For synthetic data, we used the results from OR-Tools as best-known solutions. 
\begin{table*}[ht]
\setlength{\tabcolsep}{4pt}
\centering
\caption{Ablation Studies}
\label{tab:Ablation}
\fontsize{9}{9}\selectfont
\begin{tabular}{l|ccc|ccc|ccc|ccc}
\toprule
           & \multicolumn{3}{c|}{FJSP 10x5}            & \multicolumn{3}{c|}{FJSP 20x5}            & \multicolumn{3}{c|}{FJSP 15x10}           & \multicolumn{3}{c}{FJSP 20x10}            \\
           & Obj             & Gap             & Time(s) & Obj             & Gap             & Time(s) & Obj             & Gap             & Time(s) & Obj             & Gap              & Time(s) \\
\midrule
DAN      & 108.76          & 12.96\%         & 0.32    & 197.03          & 4.72\%          & 0.63    & 160.61          & 11.9\%          & 0.97    & 199.73          & 1.95\%           & 1.31    \\
\midrule
DME    & 106.17          & 10.24\%         & 0.15 & 193.36          & 2.8\%           & 0.30 & 157.91          & 10.05\%         & 0.46 & \textbf{193.65} & \textbf{-1.17\%} & 0.62 \\
DM2E  & \textbf{105.12} & \textbf{9.13\%} & 0.19 & 193.88          & 3.07\%          & 0.38 & 159.06          & 10.84\%         & 0.58 & 194.29          & -0.84\%          & 0.80 \\
DM3E  & 105.47          & 9.51\%          & 0.23 & 194.11          & 3.19\%          & 0.46 & 158.11          & 10.19\%         & 0.69 & 194.2           & -0.89\%          & 0.97 \\
\midrule
CA       & 105.97          & 10.0\%          & 0.14 & 193.15          & 2.67\%          & 0.31 & 158.44          & 10.45\%         & 0.50 & 194.67          & -0.65\%          & 0.61 \\
M-CA & 105.8           & 9.85\%          & 0.19 & \textbf{193.12} & \textbf{2.66\%} & 0.38 & \textbf{157.56} & \textbf{9.78\%} & 0.58 & 193.9           & -1.03\%          & 0.80 \\

M2-CA & 105.7  & 9.82\% & 0.22 & 194.5  & 3.4\%  & 0.43 & 158.51 & 10.46\% & 0.67 & 196.7  & 0.39\%  & 0.92 \\
M3-CA & 105.78 & 9.83\% & 0.26 & 194.19 & 3.21\% & 0.51 & 162.3  & 13.14\% & 0.77 & 195.27 & -0.35\% & 1.07 \\
\bottomrule
\end{tabular}
\end{table*}
We report results for greedy strategies (i.e., -G) and sampling strategies (i.e., -S). The greedy strategy has only one trajectory and selects the operation-machine pair with the highest probability at each step. The sampling strategy parallelizes 100 trajectories for an instance and samples each step of the selection by probability, and finally selects the best solution among all trajectories. The sampling strategy ensures adequate exploration of model performance and improves the quality of solutions, but increases time and memory overhead. However, this time increase does not come from the model's inference process, but from autoregressive interactions with the environment, which will be further analyzed in the Appendix \ref{app:time}. 
As shown in Table \ref{tab:results3}, our encoder-decoder approach (i.e., M-CA) achieves the best results overall, with sampling strategy results that outperform all learning-based approaches as well as priority dispatch rules. Our encoder-only approach (i.e., DME) is also competitive and achieves the best greedy results on Hurink(vdata). For the greedy strategy, in Figure \ref{Benchmarks}, DAN achieved the best results on Hurink's edata, and MLP was best on the Barnes dataset. Our approaches achieve the first or second minimum makespan on all benchmarks. 
In addition, our Mamba-based approaches have great potential to solve JSSP instances, outperforming the L2D \cite{zhang2020learning} approach on various benchmarks, which is specifically designed for JSSP, as we will illustrate in Appendix \ref{app:jsp}.

\noindent\textbf{Results on synthetic data.} Second, we test the performance of our method on synthetic data. 
As in the previous approach \cite{song2023flexible}, for problem sizes of $10\times5$, $20\times5$, $15\times10$ and $20\times10$, we train the model on each size and test it on the same size. For the large-scale problems, $30\times10$ and above, we use the model trained on $20\times10$ for generalization testing. As shown in Table \ref{tab:results1}, our Mamba-based approach outperforms the previous learning-based approaches and has a time advantage. Both our encoder-only (i.e., DME) and encoder-decoder (i.e., M-CA) methods outperform the previous state-of-the-art method, DAN, and beat the exact solver OR-tools on $20\times10$. Further, our proposed efficient cross-attention decoder can improve the performance of the DME and maintain less time overhead. Compared to the previous DAN approach, our methods have speedups of $2\times$ and $1.6\times$. Although the PDRs are able to obtain solutions in a shorter time, their short-sighted nature leads to suboptimal solutions.

In Table \ref{tab:results2}, we consider manufacturing scenarios in real factories where the number of jobs is much larger than the number of machines. For large-scale problems ($30\times10$ and $40\times10$) as well as super-sized problems ($100\times10$ and above), our approaches show great generalization. At these sizes, we test with the greedy strategy due to the longer solution time. However, the previous HGNN method was unable to solve the $10000\times10$ problem due to its complex graph structure, which exceeded the GPU memory of our device (24GB). And OR-Tools solver exceeded the system memory limit of our device (128 GB). On super-sized problems larger than $100\times10$, OR-tools is unable to provide exact solutions. On $1000\times10$, we extended its time limit to one hour and still could not obtain high-quality solutions. As a comparison, our M-CA approach maintains high performance and achieves significantly better results than OR-Tools and PDRs. Thus, our approach has the ability to solve large-scale scheduling problems in real manufacturing scenarios. More results and analysis are presented in Appendix \ref{app:super-size}.

\vspace{-10pt}
\subsection{Ablation Studies}
\noindent\textbf{On Architecture of Mamba Network.} Similar to the attention model, multiple Mamba blocks can be concatenated to form larger models. We attempted to use encoders with multilayer Mamba blocks for feature extraction. As shown in Table \ref{tab:Ablation}, we trained and tested the Dual Mamba Encoder of 1-3 layers of Mamba blocks on synthetic data of different sizes. The setup using two layers of Mamba blocks (i.e. DM2E) performs best on size $10\times5$ instances, but is no longer advantageous as the problem size increases. Therefore, the encoders using only one layer of Mamba block are simple but efficient, and offer strengths in inference speed.

\noindent\textbf{On Impact of Cross Attention.} Our proposed efficient cross-attention based decoder can be used alone (i.e., CA in Table \ref{tab:Ablation}). The CA mechanism performs well on small-scale problem sizes $10\times5$ and $20\times5$, but its performance degrades on large-scale problems $15\times10$ and $20\times10$. Therefore, this performance degradation can be mitigated by combining the Mamba encoder and CA decoder. Finally, our encoder-decoder approach (i.e., M-CA) achieves better results on all 4 problem sizes. We further tested combining the CA decoder with the multilayer Mamba encoder (i.e., M2-CA and M3-CA), and the experiment results show that the multilayer Mamba encoder does not significantly improve the performance. Therefore, we finally used a single-layer encoder.

\vspace{-10pt}
\section{Conclusion}
In this study, we introduce Mamba-CrossAttention, a novel architecture leveraging Mamba for comprehensive sequence modelling in FJSP. Unlike prevalent graph-attention-based frameworks known for their computational intensity in FJSP, our proposed model demonstrates superior efficiency. Experimental results showcase that our method achieves faster solving speeds and outperforms state-of-the-art learning-based approaches for FJSP across diverse benchmark scenarios. This highlights the efficacy and practicality of our proposed architecture in addressing the complexities of the Flexible Job Shop Problem while enhancing computational efficiency and performance outcomes. We will further optimize the model and environment to provide even faster end-to-end solution speed to scheduling problems, exploiting the potential for real-time scheduling in manufacturing scenarios.

\bibliographystyle{ACM-Reference-Format}
\bibliography{example_paper}

\appendix

\section{Details of raw features}   \label{app:raw-feature}
For operations, machines, and operation-machine pairs, we used the same raw features as in the previous work DAN. Their details are as follows.

\textit{1) Features of operations:} For each operation $O_{ij}\in \mathcal{O}(t)$ ,the feature vector $h_{O_{ij}}$ contains 10 elements:
\begin{itemize}
       \item Scheduling flag: indicates whether operation $O_{ij}$ is scheduled.
	\item Minimum processing time of $O_{ij}$ among all machines.
	\item Average processing time of $O_{ij}$ among all machines.
	\item Span of processing time of $O_{ij}$ among all machines.
	\item Proportion of machines that $O_{ij}$ can be processed on.
	
	\item Estimated Lower bound of the completion time: $\underline{C}(O_{ij}) = \underline{C}(O_{i(j-1)})+\min \limits_{k \in M_{ij}} p_{ij}^{k}$, which represents the sum of the lower bound of the completion time of the immediate precedence operation of $O_{ij}$ and the minimum processing time of $O_{ij}$ on all eligible machines.
	\item Remaining number of operations: the number of unscheduled operations in job $J_i$.
	\item Remaining workload: the sum of average processing time of unscheduled operations in job $J_i$.
	\item Waiting time: the time from when the operation is eligible to the current time $Ts$, which is 0 when the operation is not eligible.
	\item Remaining processing time: If operation $O_{ij}$ is being processed, the time from $T_s$ until the completion of processing.
\end{itemize}

\textit{2) Features of machines:} For each machine $M_k \in \mathcal{M}(t)$, the feature vector $h_{M_k}$ contains 8 elements:
\begin{itemize}
       \item Working flag: indicates whether the machine $M_k$ is processing an operation.
	\item Minimum processing time among all operations.
	\item Average processing time of operations that $M_k$ can process.
	\item Number of unscheduled operations that $M_k$ can process.
	\item Number of candidates that $M_k$ can process.
	\item Free time: the moment when machine $M_k$ stops working.
	\item Waiting time: the time from the free time until $T_s$, indicating the time to wait for the operation to be processed.
	\item Remaining processing time: the time from $T_s$ until the free time, indicating the time to complete the processing operation.
\end{itemize}

\textit{3) Features of eligible operation-machine pairs:}  For each eligible operation-machine pair $(O_{ij},M_k) \in \mathcal{A}(t)$, the feature vector $h_{(O_{ij},M_k)}$ contains 8 elements:
\begin{itemize}
	\item Processing time $p_{ij}^{k}$ of $O_{ij}$ on $M_k$.
	\item Ratio of $p_{ij}^{k}$ to the maximum processing time of $O_{ij}$.
	\item Ratio of $p_{ij}^{k}$ to the maximum processing time of candidates that $M_k$ can process.
	\item Ratio of $p_{ij}^{k}$ to the maximum processing time of unscheduled operations.
	\item Ratio of $p_{ij}^{k}$ to the maximum processing time of unscheduled operations that $M_k$ can process.
	\item Ratio of $p_{ij}^{k}$ to the maximum processing time of eligible pairs.
	\item Ratio of $p_{ij}^{k}$ to remaining workload of $J_i$.
	\item Sum of waiting time of $O_{ij}$ and $M_k$.
\end{itemize} 

When the state is transitioned to the next state, all features are changed according to the environment. Specifically, the operation's scheduling flags, wait times, and the machine's work flags, free time, etc. are recalculated based on the change of environment, and the details of the update rules will be explained in the code.

\section{Details of the Training Algorithm} \label{app:Algorithm}
In Algorithm \ref{the_algorithm}, we use pseudo-code to describe our training process.

\begin{algorithm}[ht]
	\caption{Training Mamba for FJSP}
	\label{the_algorithm}
	\begin{algorithmic}[1]
		\STATE {\bfseries Input:}
		A Dual Mamba Encoder or M-CA Encoder-Decoder with initial parameters $\Theta = \{\omega, \theta,\phi\}$, behavior actor network $\theta_{\rm old}$, pre-sampled training environment $E$, and fixed validation environment $E_v$;
		\FOR {$n_{\rm ep} =1,2,...,N$}
		\STATE $\theta_{\rm old} \gets \theta$
		\FOR {$i =1,2,...,B$}
		
		\FOR {$t =0, 1,..., T$}
		\STATE Sample $a_{i,t}$ using $\pi_{\theta_{\rm old}}(\cdot \mid s_{i,t})$;  
		\STATE Receive the reward $r_{i,t}$ and the next state $s_{i, t+1}$;
		\STATE Update $s_{i,t} \gets s_{i,t+1}$;
		\ENDFOR
		
		\STATE Compute the generalized advantage estimates $\hat{A}_{i,t}$ for $t =0, 1,..., T$ using collected transitions;

		\ENDFOR
		
		\FOR{$k = 1,2,...,K $}
		\STATE Compute the total loss $\sum_{i=1}^{|E|}\ell_i^{\rm PPO}(\Theta)$; 
		\STATE Update all parameters $\Theta$;
		\ENDFOR

		\IF{\textit{Every} $N_{res}$ \textit{episodes}}
		\STATE Resample $B$ instances to reset the training environment;
		\ENDIF
		
		\IF{\textit{Every} $N_{\rm val}$ \textit{episodes}}
		\STATE Validate the model on $E_v$ and save the best $\Theta$;
		\ENDIF
		
		\ENDFOR
		\vspace{-0pt}
	\end{algorithmic}
\end{algorithm}

\section{Details of Benchmarks}  \label{app:benchmarks-detail}
In Table \ref{tab:bench-detail}, we show the public benchmarks we use in terms of data source, range of problem sizes, range of operation processing times and number of instances.

\begin{table}[!h]
\setlength{\tabcolsep}{3pt}
\centering
\caption{Details of Benchmarks}
\label{tab:bench-detail}
\begin{tabular}{l|c|c|c|c}
\toprule
Benchmarks    & Source & Sizes       & Range       & Instances \\
\midrule
Brandimarte   &  \cite{brandimarte1993routing}      & 10x6-20x15  & {[}1,20{]}  & 10        \\
Hurink(rdata) & \cite{hurink1994tabu}       & 10x5-30x10  & {[}1,99{]}  & 40        \\
Hurink(edata) &  \cite{hurink1994tabu}      & 10x5-30x10  & {[}1,99{]}  & 40        \\
Hurink(vdata) & \cite{hurink1994tabu}       & 10x5-30x10  & {[}1,99{]}  & 40        \\
Barnes        & \cite{barnes1996flexible}       & 10x11-15x17 & {[}1,99{]}  & 21        \\
Dauzere       &  \cite{dauzere1997integrated}      & 10x5-20x10  & {[}1,100{]} & 18       \\
\bottomrule
\end{tabular}

\end{table}

\section{Details of the Baselines}  \label{app:pdr}
In this solution, we represent the scheduling process of baseline PDRs in detail, explaining how they solve FJSP instances.

\begin{itemize}
    \item First In First Out (FIFO): selects the earliest ready candidate operation and the earliest ready compatible machine.
    \item Most Operations Remaining (MOPNR): selects the candidate operation with the most remaining successor operations and the machine that can process that operation immediately.
 \item Shortest Processing Time (SPT): selects the compatible operation-machine pair with the shortest processing time.
 \item Most Work Remaining (MWKR): selects the candidate operation with the most average processing time remaining for the successor and the machine that can process that operation immediately.
\end{itemize}


\begin{table*}[!ht]
\setlength{\tabcolsep}{3pt}
\centering
\caption{Composition of solution time}
\label{tab:time}
\fontsize{9}{10}\selectfont
\begin{tabular}{l|cccc|cccc|cccc}
\toprule
        & \multicolumn{4}{c|}{FJSP 20x10}          & \multicolumn{4}{c|}{FJSP 30x10}          & \multicolumn{4}{c}{FJSP 40x10}          \\
        & Obj    & Model(s) & Other(s) & Total(s) & Obj    & Model(s) & Other(s) & Total(s) & Obj    & Model(s) & Other(s) & Total(s) \\
\midrule
DAN-G   & 199.73 & 1.06     & 0.39     & 1.45     & 283.57 & 1.63     & 0.62     & 2.24     & 373.22 & 2.02     & 0.78     & 2.80     \\
DAN-S   & 195.14 & 1.09     & 3.88     & 4.97     & 280.51 & 1.74     & 8.48     & 10.22    & 370.85 & 2.28     & 14.04    & 16.32    \\
\midrule
DME-G & 193.65 & 0.20     & 0.41     & 0.61     & 277.17 & 0.30     & 0.63     & 0.93     & 366.44 & 0.40     & 0.89     & 1.28     \\
DME-S & 190.13 & 0.27     & 3.99     & 4.26     & 274.48 & 0.39     & 7.99     & 8.38     & 364.21 & 0.59     & 14.32    & 14.91    \\
M-CA-G  & 193.9  & 0.36     & 0.42     & 0.78     & 277    & 0.53     & 0.64     & 1.17     & 367.06 & 0.71     & 0.88     & 1.59     \\
M-CA-S  & 189.83 & 0.50     & 4.11     & 4.61     & 274.78 & 0.76     & 8.90     & 9.66     & 364.94 & 1.02     & 15.16    & 16.18    \\
\bottomrule
\end{tabular}

\end{table*}
\section{Analysis of the solution time} \label{app:time}
As shown in Table \ref{tab:time}, we report the composition of solution time, including the model's feature extraction time (i.e., Model), other time (i.e., Other, which mainly involves interaction with the environment), and end-to-end solution generation time (i.e., Total). On larger size instances (e.g., $20\times10$, $30\times10$, and $40\times10$), the increase in solution time stems primarily from multiple autoregressive interactions with the environment. We find that the sampling strategy significantly increases the time except the model inference time, which is related to the parallel processing of multiple trajectories in the environment. The sampling strategy has a small effect on the feature extraction time. For a fair comparison, our approaches use the same environment as DAN. Compared to DAN, our approaches have a faster feature extraction speed, resulting in a shorter solution generation time. We will further optimize the environment settings to speed up the sampling strategy.

\section{Results on JSSP benchmarks}  \label{app:jsp}

\begin{table*}[!ht]
\setlength{\tabcolsep}{3pt}
\centering
\caption{Results on JSSP benchmarks}
\label{tab:results_jsp}
\fontsize{9}{10}\selectfont
\begin{tabular}{l|c|c|c|c|c|c|c|c|c}
\toprule
        &      & Ta 15x15         & Ta 20x15         & Ta 20x20         & Ta 30x15         & Ta 30x20         & Ta 50x15         & Ta 50x20         & Ta 100x20       \\
\midrule
L2D     & Obj  & 1547.4           & 1774.7           & 2128.1           & 2378.8           & 2603.9           & 3393.8           & 3593.9           & 6097.6          \\
        & Gap  & 26.00\%          & 30.00\%          & 31.60\%          & 33.00\%          & 33.60\%          & 22.40\%          & 26.50\%          & 13.60\%         \\

\midrule
DAN-G   & Obj  & 1425.1           & 1634.4           & 1916.6           & 2132.8           & 2355.6           & 3136.6           & 3244.4           & 5708.5          \\
        & Gap  & 15.97\%          & 19.72\%          & 18.49\%          & 19.23\%          & 20.94\%          & 13.13\%          & 14.11\%          & 6.39\%          \\
        & Time & 1.49             & 2.00             & 2.72             & 3.08             & 4.19             & 5.34             & 7.52             & 17.12           \\
DAN-S   & Obj  & 1350.6           & 1553.8           & 1816.6           & 2035.5           & 2278.5           & 3025.5           & 3148.7           & 5583.4          \\
        & Gap  & 9.91\%           & 13.84\%          & 12.32\%          & 13.81\%          & 16.96\%          & 9.12\%           & 10.72\%          & 4.06\%          \\
        & Time & 6.97             & 11.70            & 35.09            & 32.33            & 75.68            & 87.95            & 219.33           & 883.28          \\
\midrule
DME-G & Obj  & \textbf{1414.7}  & 1617             & 1915.2           & 2149.6           & 2348.2           & 3111.7           & 3196.8           & \textbf{5606.6} \\
        & Gap  & \textbf{15.16\%} & 18.48\%          & 18.44\%          & 20.24\%          & 20.53\%          & 12.25\%          & 12.42\%          & \textbf{4.49\%} \\
        & Time & 0.77             & 1.07             & 1.55             & 1.70             & 2.43             & 2.98             & 4.51             & 10.45           \\
DME-S & Obj  & \textbf{1349.4}  & 1524.8           & \textbf{1802.5}  & 2035.8           & 2247.1           & 2995.9           & 3115             & \textbf{5469.2} \\
        & Gap  & \textbf{9.81\%}  & 11.72\%          & \textbf{11.46\%} & 13.84\%          & 15.36\%          & 8.05\%           & 9.55\%           & \textbf{1.93\%} \\
        & Time & 6.92             & 11.77            & 35.16            & 33.16            & 77.82            & 90.51            & 226.32           & 930.67          \\
M-CA-G  & Obj  & 1432.3           & \textbf{1603.3}  & \textbf{1904.8}  & \textbf{2128.8}  & \textbf{2341}    & \textbf{3109.7}  & \textbf{3193.9}  & 5610.1          \\
        & Gap  & 16.54\%          & \textbf{17.50\%} & \textbf{17.80\%} & \textbf{19.04\%} & \textbf{20.18\%} & \textbf{12.18\%} & \textbf{12.32\%} & 4.56\%          \\
        & Time & 0.93             & 1.25             & 1.75             & 1.92             & 2.72             & 3.42             & 4.99             & 12.10           \\
M-CA-S  & Obj  & 1356.6           & \textbf{1519.4}  & 1811.9           & \textbf{2018.9}  & \textbf{2242.6}  & \textbf{2976}    & \textbf{3091.1}  & 5471.1          \\
        & Gap  & 10.39\%          & \textbf{11.34\%} & 12.05\%          & \textbf{12.91\%} & \textbf{15.12\%} & \textbf{7.35\%}  & \textbf{8.71\%}  & 1.97\%          \\
        & Time & 6.61             & 11.53            & 35.28            & 33.11            & 77.97            & 89.68            & 229.06           & 905.95         \\

\midrule

        &      & DMU 20x15        & DMU 20x20        & DMU 30x15        & DMU 30x20        & DMU 40x15        & DMU 40x20        & DMU 50x15        & DMU 50x20        \\
\midrule
L2D & Obj & 4215.3  & 4804.5  & 5557.9 & 5967.4  & 6663.9  & 7375.8  & 8179.4  & 8751.6  \\
    & Gap & 38.95\% & 37.74\% & 41.86\%                        & 39.48\% & 35.39\% & 39.37\% & 36.20\% & 38.85\%\\
\midrule
DAN-G   & Obj  & 3769.6           & 4219.5           & 5002.1           & 5567             & 6161.6           & 6791             & 7521             & 8123.2           \\
        & Gap  & 24.13\%          & 21.08\%          & 27.82\%          & 30.05\%          & 25.24\%          & 28.27\%          & 25.22\%          & 28.62\%          \\
        & Time & 2.29             & 3.04             & 3.28             & 3.76             & 4.59             & 6.67             & 4.73             & 8.32             \\
DAN-S   & Obj  & 3559.9           & 4009.8           & 4826.3           & 5250.2           & 5932.2           & 6547.9           & 7324             & 7924.5           \\
        & Gap  & 17.28\%          & 15.18\%          & 23.34\%          & 22.54\%          & 20.52\%          & 23.70\%          & 21.95\%          & 25.46\%          \\
        & Time & 12.79            & 36.42            & 35.52            & 64.43            & 65.27            & 142.14           & 72.70            & 234.09           \\
\midrule
DME-G & Obj  & 3722             & 4166.5           & 4844.5           & 5267.6           & \textbf{5866.6}  & \textbf{6542.5}  & \textbf{6964.3}  & \textbf{7660.5}  \\
        & Gap  & 22.56\%          & 19.65\%          & 24.20\%          & 23.23\%          & \textbf{19.41\%} & \textbf{23.86\%} & \textbf{16.30\%} & \textbf{21.76\%} \\
        & Time & 1.10             & 1.51             & 1.72             & 2.46             & 2.44             & 3.47             & 2.65             & 4.04             \\
DME-S & Obj  & 3473.7           & 4004.8           & \textbf{4573}    & 5046.3           & \textbf{5558.9}  & \textbf{6269.6}  & \textbf{6679.5}  & \textbf{7363.9}  \\
        & Gap  & 14.55\%          & 14.96\%          & \textbf{17.35\%} & 18.15\%          & \textbf{13.36\%} & \textbf{18.70\%} & \textbf{11.60\%} & \textbf{17.13\%} \\
        & Time & 11.89            & 35.15            & 33.98            & 77.83            & 59.39            & 143.75           & 86.15            & 224.31           \\
M-CA-G  & Obj  & \textbf{3590.4}  & \textbf{4151.3}  & \textbf{4811.5}  & \textbf{5244.8}  & 5927.4           & 6544.2           & 7107.3           & 7779.5           \\
        & Gap  & \textbf{18.53\%} & \textbf{19.25\%} & \textbf{23.33\%} & \textbf{22.78\%} & 20.68\%          & 23.90\%          & 18.61\%          & 23.49\%          \\
        & Time & 1.40             & 2.00             & 2.14             & 3.05             & 2.93             & 4.17             & 3.29             & 5.11             \\
M-CA-S  & Obj  & \textbf{3452.8}  & \textbf{3984.6}  & 4591.3           & \textbf{5030.3}  & 5600             & 6290.3           & 6794.1           & 7421.3           \\
        & Gap  & \textbf{13.90\%} & \textbf{14.50\%} & 17.71\%          & \textbf{17.67\%} & 14.09\%          & 19.01\%          & 13.45\%          & 17.86\%          \\
        & Time & 12.44            & 37.55            & 35.58            & 83.52            & 62.47            & 149.74           & 94.50            & 246.25          \\

\bottomrule
\end{tabular}

\end{table*}
The job shop scheduling problem (JSSP) is a simple variant of the flexible job shop scheduling problem (FJSP) in which each operation can be processed on only one machine. To further explore the ability of our approaches to solve JSSP, we train our models and DAN on JSSP $30\times20$ instances, and test them on public benchmarks. We use 1000 training iterations, each iteration containing 20 instances, and replace them every 20 iterations. We use the validation set to validate every 10 iterations and get the best models for testing. The operation times in training and validation data are sampled from the distribution of $U (1, 99)$. 

We test the generalizability of our approaches on popular JSSP benchmarks, including Taillard's instances \cite{taillard1993benchmarks} and DMU instances \cite{demirkol1998benchmarks}. For the previous work L2D \cite{zhang2020learning}, we use the best results from their paper for comparison, which outperform PDRs. As shown in Table \ref{tab:results_jsp}, our Mamba-based approaches (i.e. DME and M-CA) show strong potential in solving JSSP, surpassing DAN and L2D. On Taillard's instances, our M-CA method generalizes well and achieves the best results on multiple sizes. On DMU instances, our encoder-only Mamba approach shows better generalization, obtaining better large-scale JSSP solutions. Notably, our approach outperforms the L2D approach which was designed to solve JSSP specifically, exhibiting excellent cross-problem generalization, and we will further explore cross-problem learning in future research.

\section{Results of super-sized instances}   \label{app:super-size}

To further explore the scalability of our approach, we test our approach on super-sized FJSP instances with the greedy strategy. We test $100\times10$, $500\times10$, $1000\times10$, $5000\times10$, and $10000\times10$ FJSP instances with models trained on size $20\times10$. The datasets are generated following the settings in \cite{song2023flexible}. For the size $100\times10$, $500\times10$ and $1000\times10$, we used 10 instances for each size. For the size $5000\times10$ and $10000\times10$, we used 5 instances for each size due to its longer solving time. We report objective values, GPU memory overhead, and time to generate solutions. For OR-tools and PDRs, they do not depend on the GPU to execute, and therefore do not have GPU memory overhead. In the learning-based approaches, we further explore the solution performance of the self-attention model. We try to replace the Mamba block with a single self-attention layer and observe its performance. As shown in Table 6, the self-attention layer is able to extract operation and machine features and outperforms Mamba on problem size of $1000\times10$, but it is unable to solve problems of size above $5000\times10$ due to its huge GPU memory overheads. HGNN also faces the problem of excessive memory overhead due to its complex graph structure. Compared to DAN, the memory overhead of our approaches are slightly increased, but our approaches are able to extract the full sequence features and bring better performance. The graph attention mechanism used by DAN can only focus on the neighboring nodes in the sequence, which leads to suboptimal solutions. At the same time, in solving super-sized problems, the exact solver OR-tools is not able to achieve high-quality solutions in a reasonable time. We extended the time limit of OR-tools to 3600s at $1000\times10$, 7200s at $5000\times10$ and 18000s at $10000\times10$. 
However, it consumes too much memory and CPU cores during the solving process, exceeding the system memory limit of our device (128 GB), so we could not obtain its results on $10000\times10$. Also, at the size of $5000\times10$, the OR-tools results are far from feasible solutions. We ran it several times and its results remain the same. Thus, it is not feasible to use OR-tools to solve super-sized problems.
As a comparison, our M-CA method exhibits surprising generalization on instances exceeding 500 times the size of the training data (i.e., $20\times10$ vs. $10000\times10$), outperforming DAN and heuristic priority dispatching rules. We will further explore the large-scale generalizability of our approaches in future studies.

\begin{table*}[ht]
\centering
\caption{Results on super-sized instances}
\label{tab:super-large}
\fontsize{9}{10}\selectfont

\begin{tabular}{l|cc|cc|cc|cc|cc}
\toprule
               & \multicolumn{2}{c|}{FJSP 100x10} & \multicolumn{2}{c|}{FJSP   500x10} & \multicolumn{2}{c|}{FJSP   1000x10} & \multicolumn{2}{c|}{FJSP 5000x10} & \multicolumn{2}{c}{FJSP   10000x10} \\
               & Obj/Mem            & Time(s)       & Obj/Mem             & Time(s)        & Obj/Mem              & Time(s)        & Obj/Mem             & Time(s)       & Obj/Mem               & Time        \\
\midrule
OR-tools       & 940.7              & 1800.00    & 5038.5              & 1801.08     & 10283.9              & 3603.00     & 579152.4            & 7214.00    & -                     & -           \\
\midrule
FIFO           & 1045.7             & 1.72       & 5233.8              & 28.91       & 10302                & 87.28       & 51411               & 2029.67    & 102534.6                & 2.72h       \\
MOR            & 1043               & 1.73       & 5227.4              & 28.96       & 10296.8              & 86.13       & 51400.8             & 2097.52    & 102542.8                & 2.85h       \\
SPT            & 1026               & 1.71       & 4911.1              & 28.91       & 9570.7               & 84.22       & 47545.4             & 2088.12    & 94728.4                 & 2.71h       \\
MWKR           & 1038.7             & 1.74       & 5221.5              & 25.46       & 10273                & 85.12       & 51406.4             & 2056.33    & 102508                & 2.76h       \\
\midrule
HGNN            & 1043.2             & 6.26       & 5221                & 144.51      & 10287.1              & 1019.86     & -                   & -          & -                     & -           \\
               & 2198MB             &            & 4226MB              &             & 10600MB              &             & OOM                 &            & OOM                   &             \\
DAN            & 924.8              & 8.69       & 4662.1              & 53.18       & 9157.8               & 140.30      & 45630.2             & 2347.30    & 93871.8                 & 3.46h       \\
               & 336MB              &            & 340MB               &             & 362MB                &             & 478MB               &            & 626MB                 &             \\
\midrule
Self-Attention & 919.1              & 3.55       & 4598.9              & 72.07       & 9072.8               & 388.56      & -                   & -          & -                     & -           \\
               & 438MB              &            & 2706MB              &             & 9500MB               &             & OOM                 &            & OOM                   &             \\
DME          & \textbf{913.9}     & 3.51       & 4714.2              & 37.78       & 9617.5               & 106.10      & 49693.6             & 2272.34    & 99804.4                & 3.10h       \\
               & 340MB              &            & 378MB               &             & 400MB                &             & 684MB               &            & 1130MB                &             \\
M-CA           & 916.8              & 4.52       & \textbf{4595.7}     & 43.29       & \textbf{9038.1}      & 120.44      & \textbf{45040.6}    & 2438.62    & \textbf{90200}        & 3.34h       \\
               & 344MB              &            & 382MB               &             & 424MB                &             & 784MB               &            & 1232MB                &            \\
\bottomrule
\end{tabular}

\begin{tablenotes} \small
 	 	\item[\textbf{2}]  "OOM" indicates that the method is out of GPU memory (24GB).
  \end{tablenotes}
\end{table*}

\section{Training Curve}
\begin{figure*}[h]
\centering
\subfloat[Problem size $10\times5$]{\includegraphics[width=0.495\textwidth]{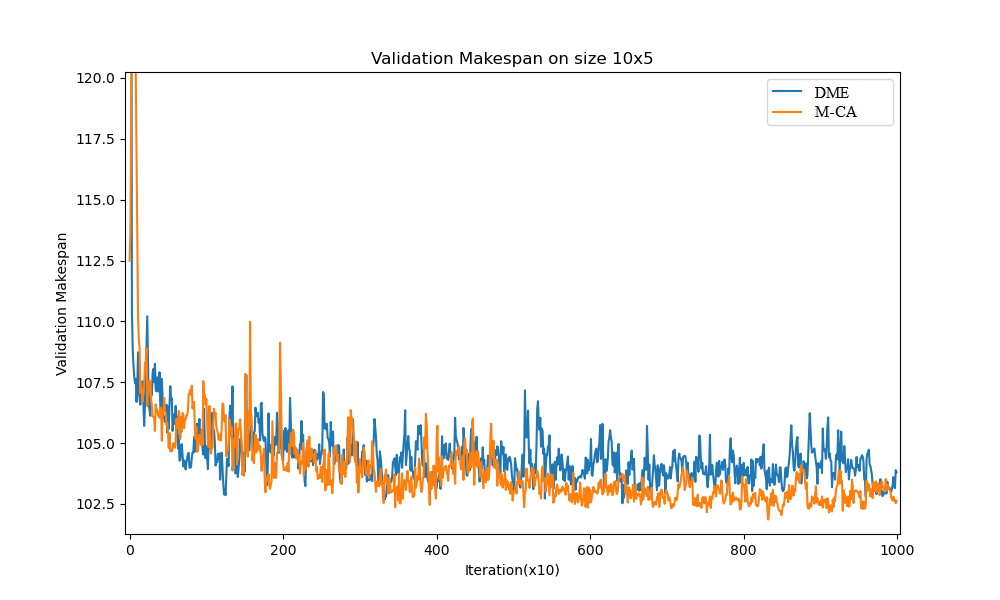}}\hspace{0.01cm}  
\subfloat[Problem size $20\times5$]{\includegraphics[width=0.495\textwidth]{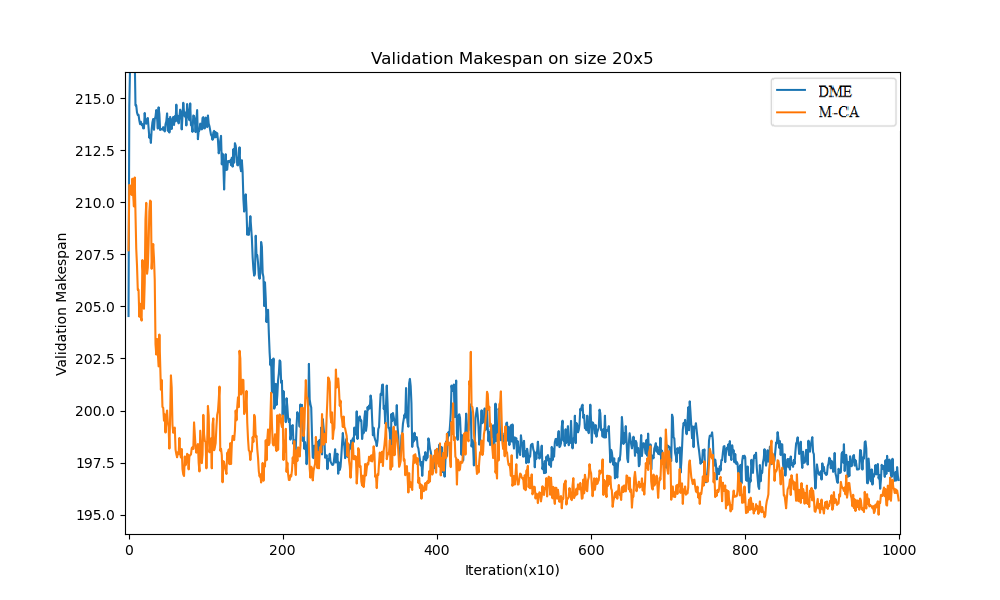}}
 \\  
\subfloat[Problem size $15\times10$]{\includegraphics[width=0.495\textwidth]{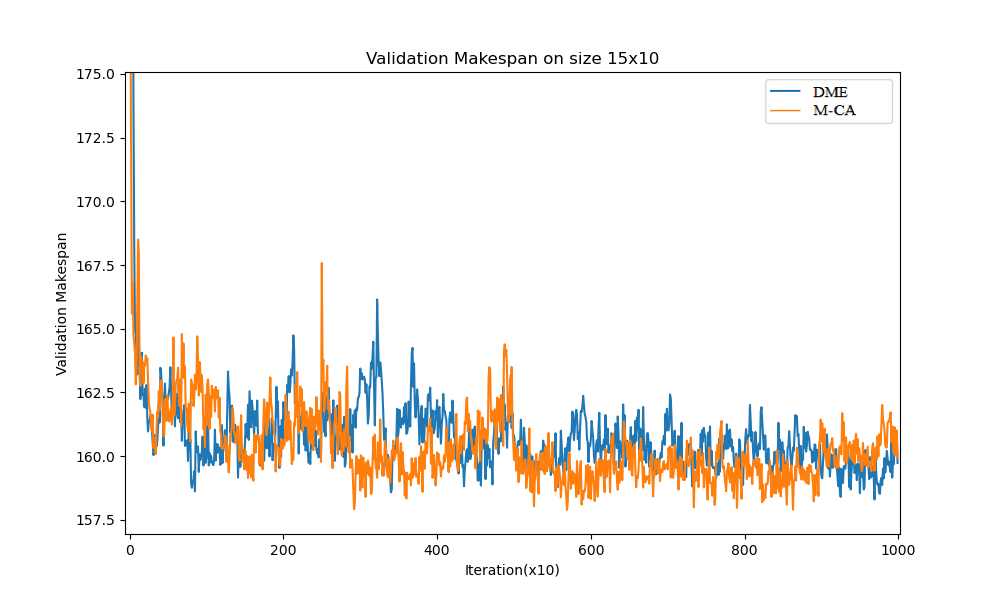}}\hspace{0.01cm}
\subfloat[Problem size $20\times10$]{\includegraphics[width=0.495\textwidth]{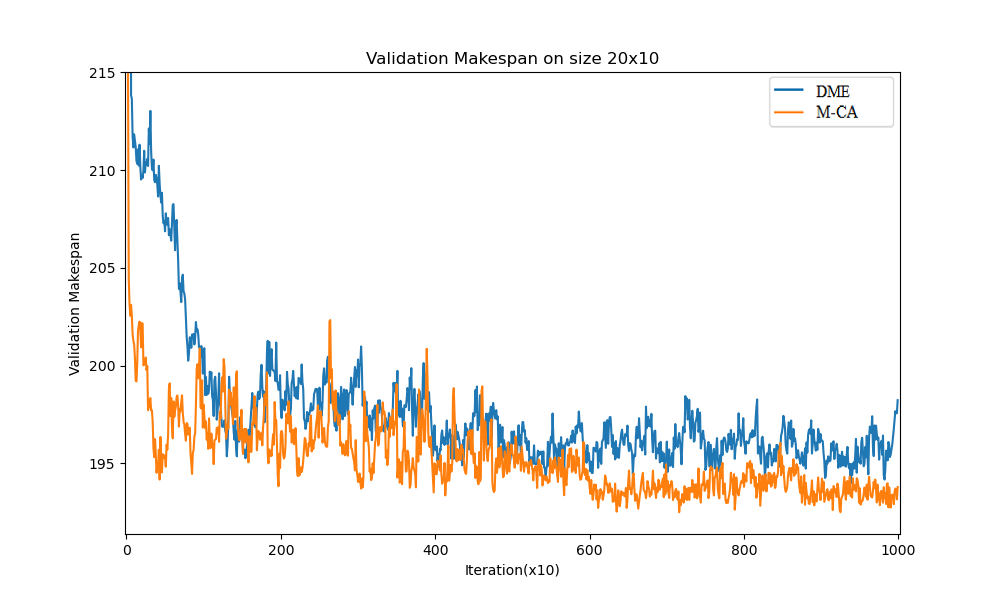}}

\caption{Training curves for all problem sizes.}  
\label{fig:curves}
\end{figure*}

In Figure \ref{fig:curves}, we show the training curves of DME and M-CA on validation datasets of 4 different training sizes ($10\times5$, $20\times5$, $15\times10$, $20\times10$).

\section{Solution Visualization}   \label{app:visualization}
\begin{figure*}[h]
\centering
\subfloat[HGNN  Makespan:259  Gap:31.47\%]{\includegraphics[width=0.495\textwidth]{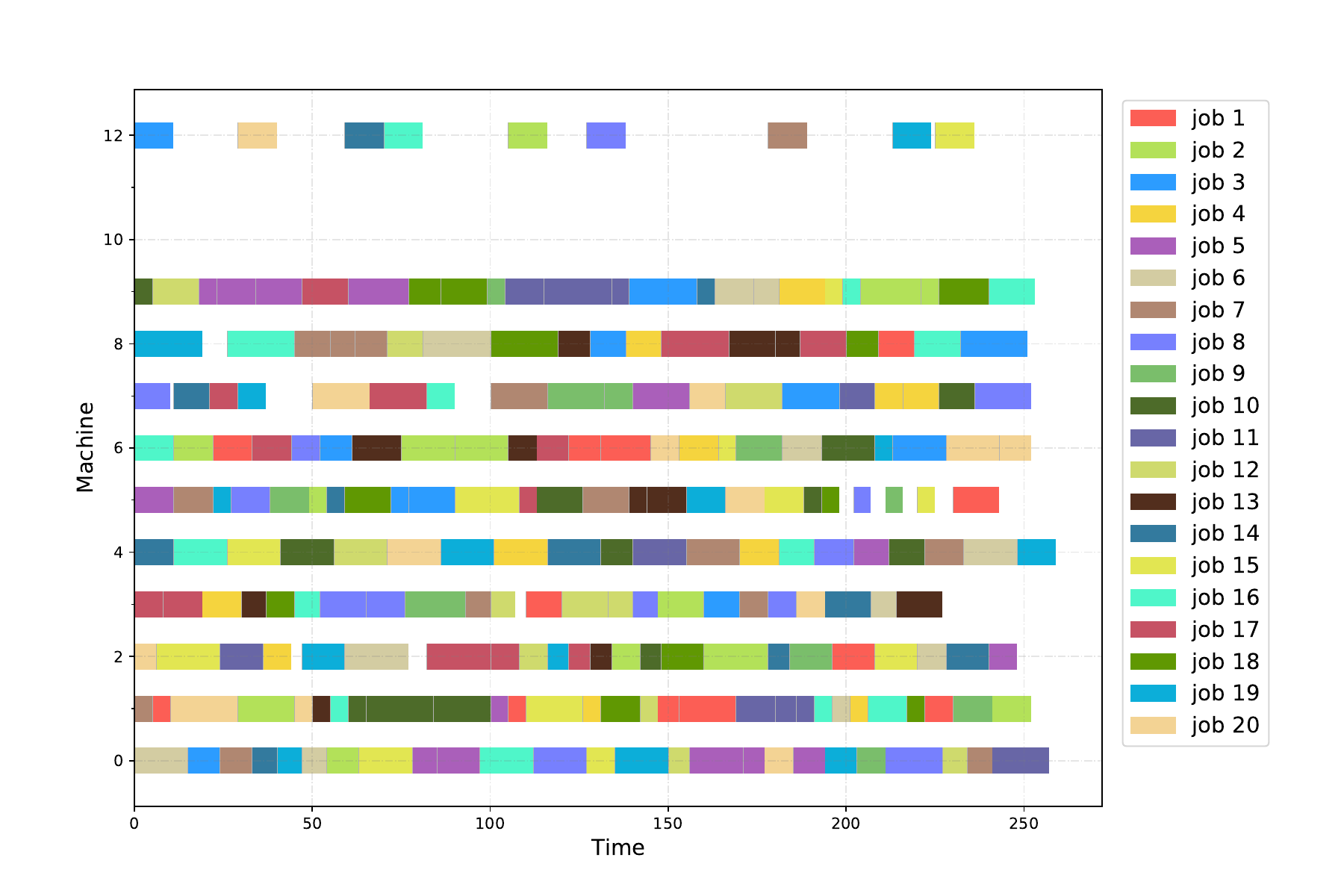}}\hspace{0.01cm}  
\subfloat[DAN  Makespan:239  Gap:21.32\%]{\includegraphics[width=0.495\textwidth]{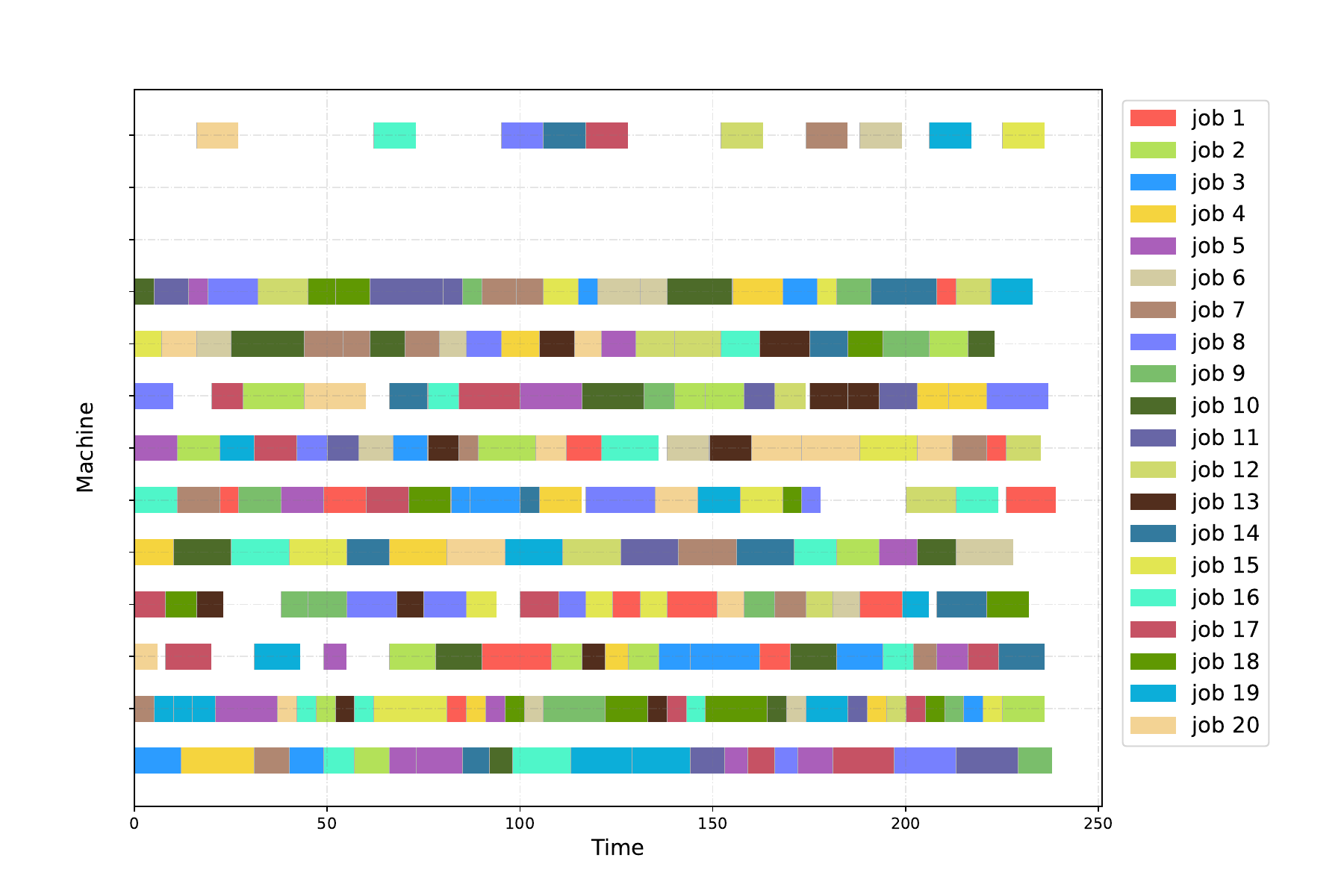}}
 \\  
\subfloat[DME  Makespan:228  Gap:15.74\%]{\includegraphics[width=0.495\textwidth]{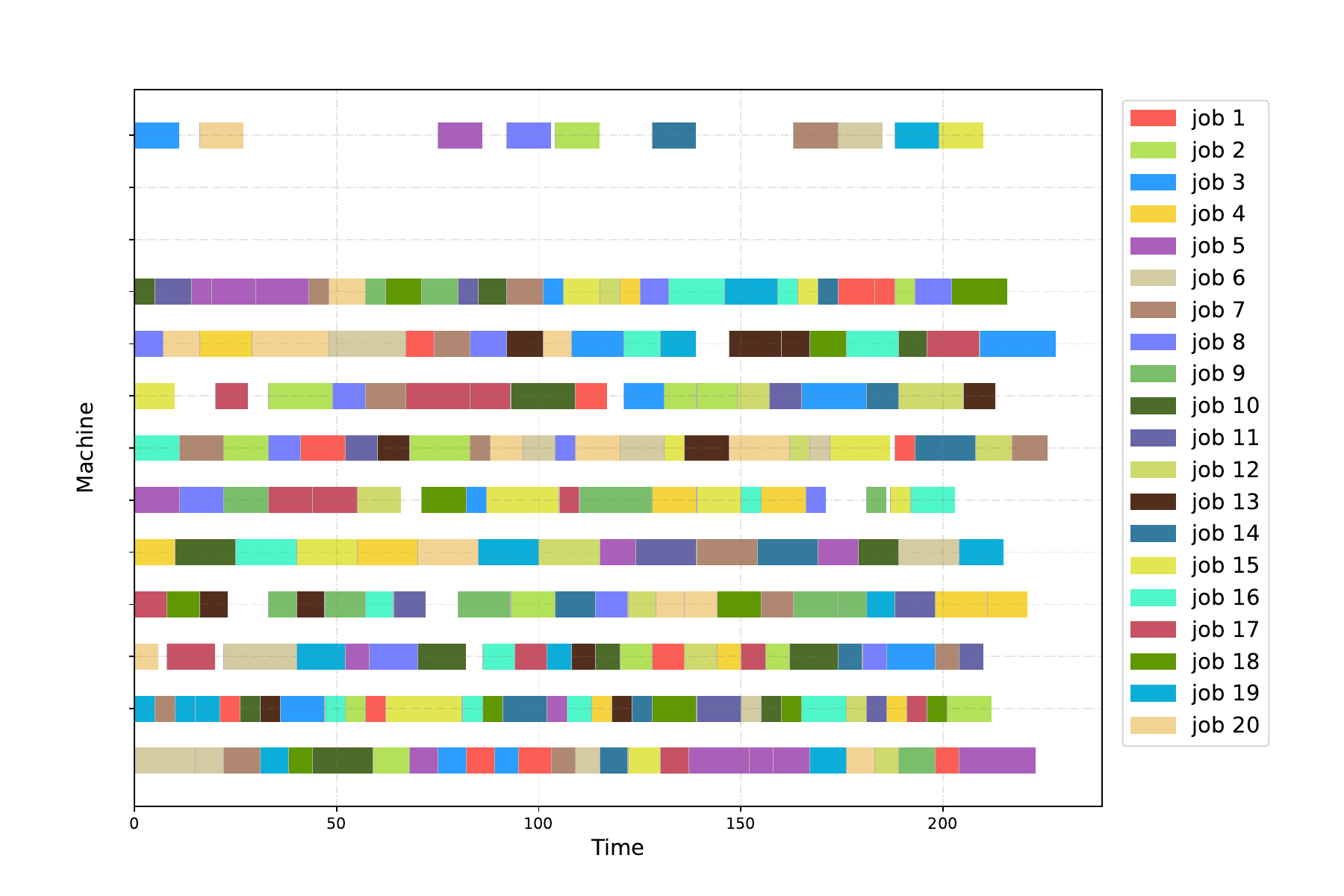}}\hspace{0.01cm}
\subfloat[M-CA  Makespan:221  Gap:12.18\%]{\includegraphics[width=0.495\textwidth]{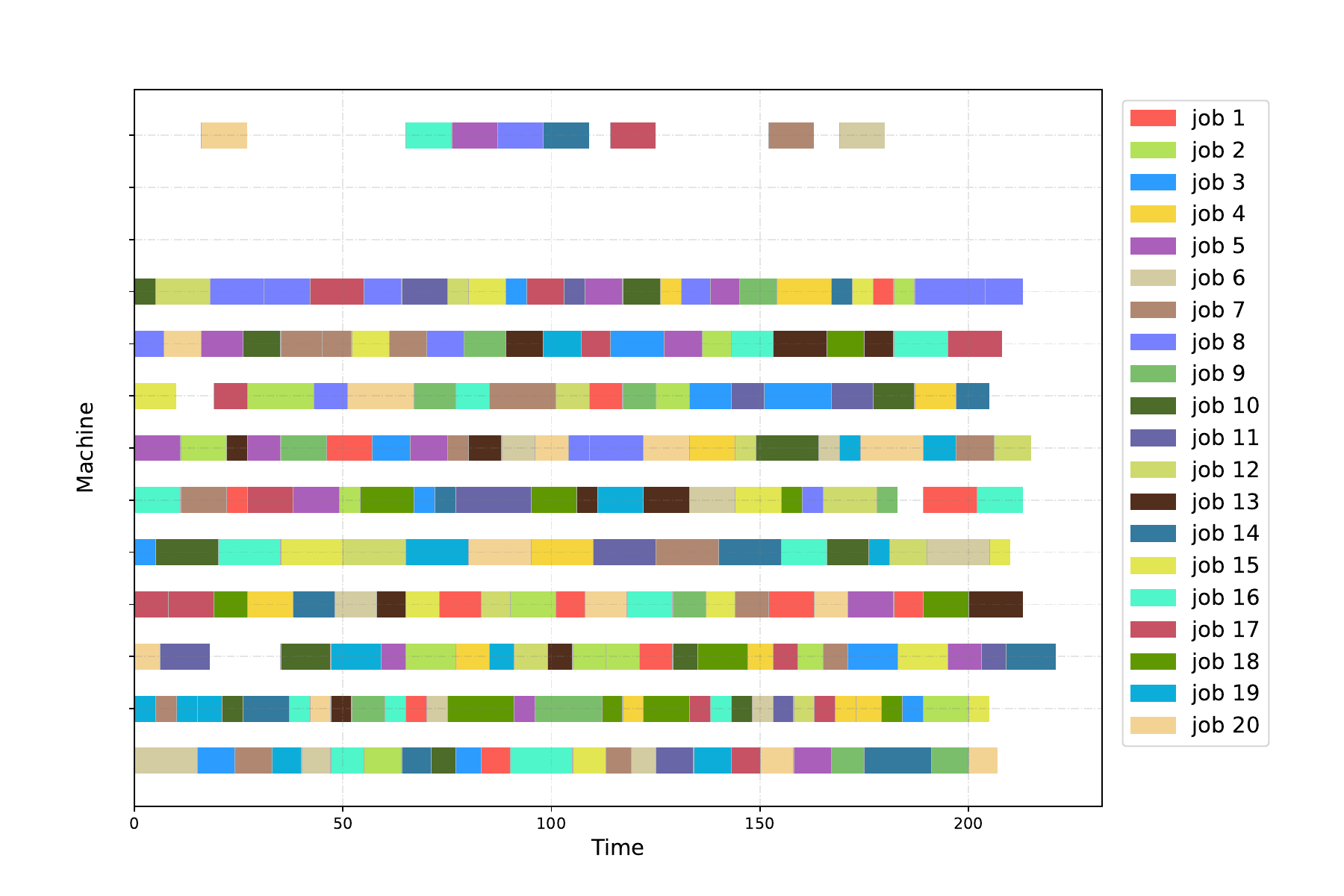}}

\caption{Brandimarte Mk10.}  
\label{fig:BrandimarteMk10}
\end{figure*}

\begin{figure*}[h]
\centering
\subfloat[HGNN  Makespan:621  Gap:17.17\%]{\includegraphics[width=0.495\textwidth]{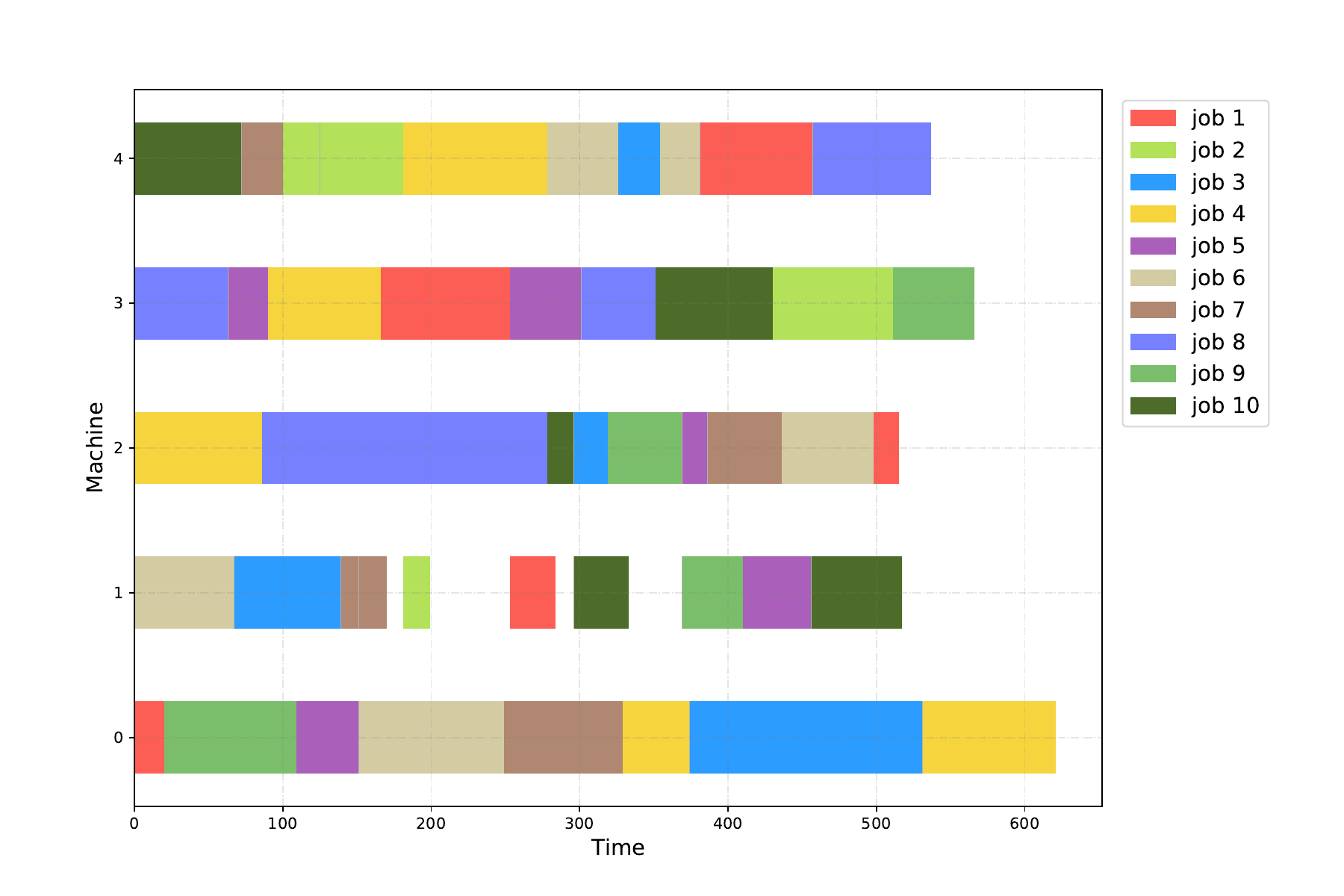}}\hspace{0.01cm}  
\subfloat[DAN  Makespan:633  Gap:19.43\%]{\includegraphics[width=0.495\textwidth]{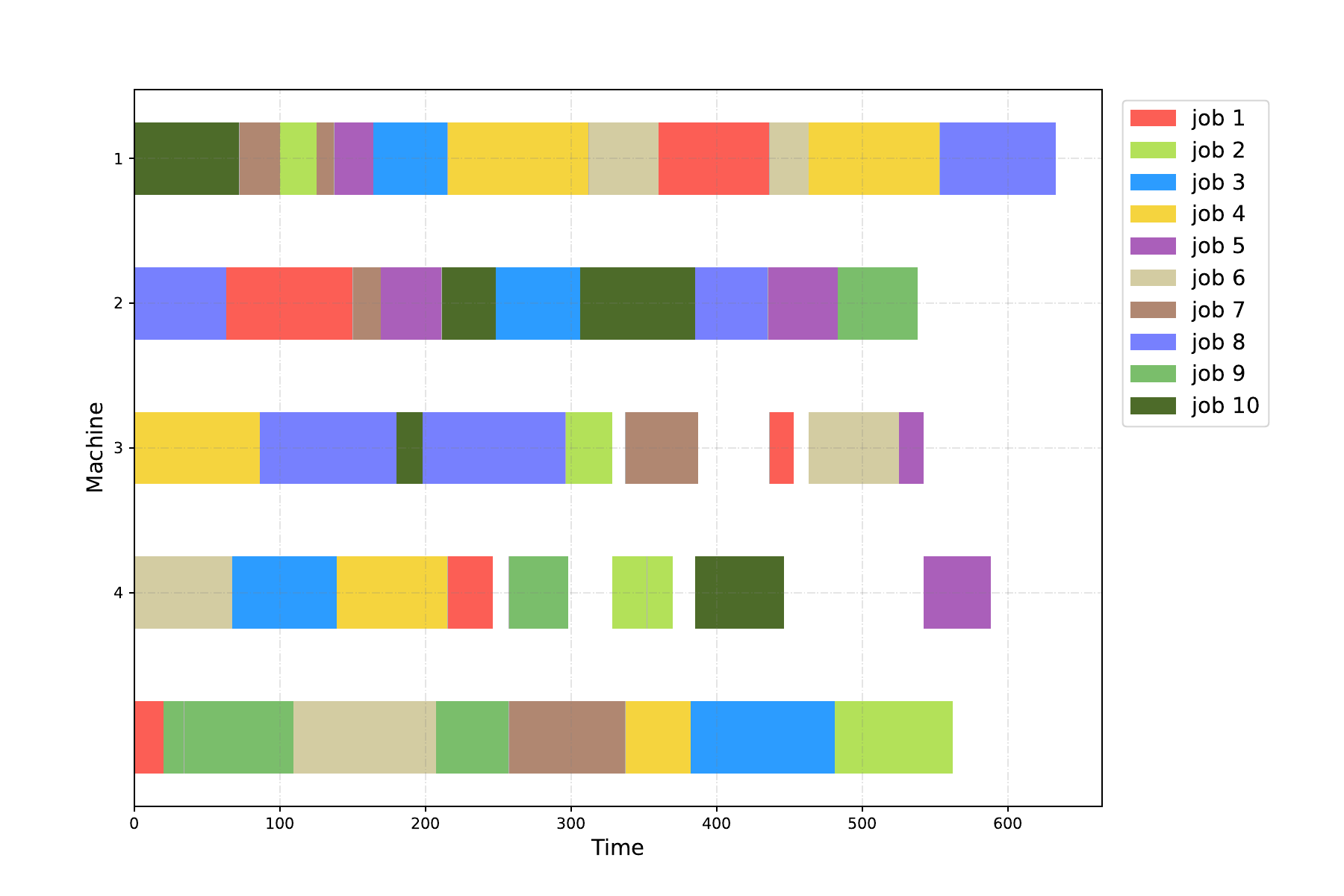}}
 \\  
\subfloat[DME  Makespan:609  Gap:14.91\%]{\includegraphics[width=0.495\textwidth]{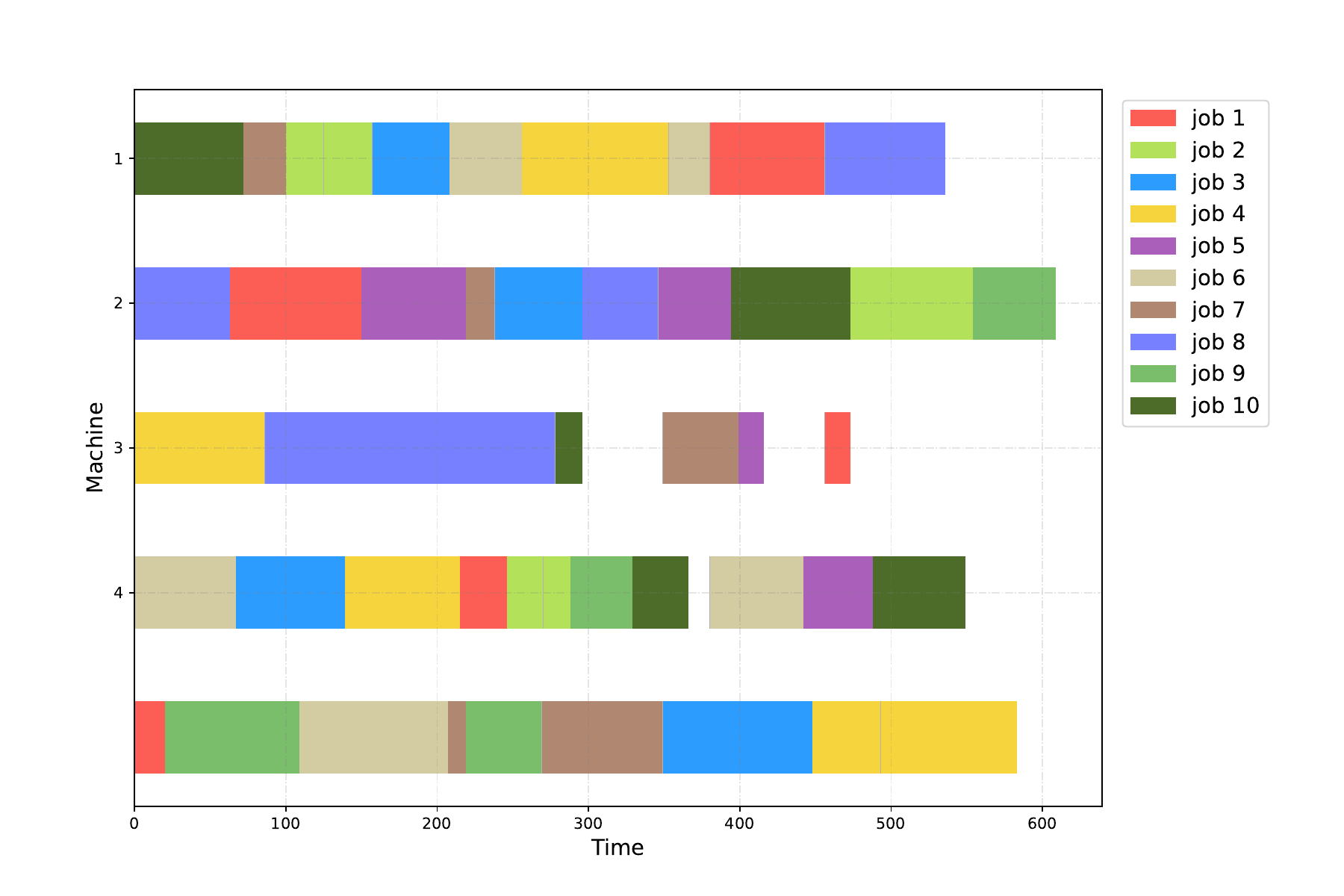}}\hspace{0.01cm}
\subfloat[M-CA  Makespan:569  Gap:7.36\%]{\includegraphics[width=0.495\textwidth]{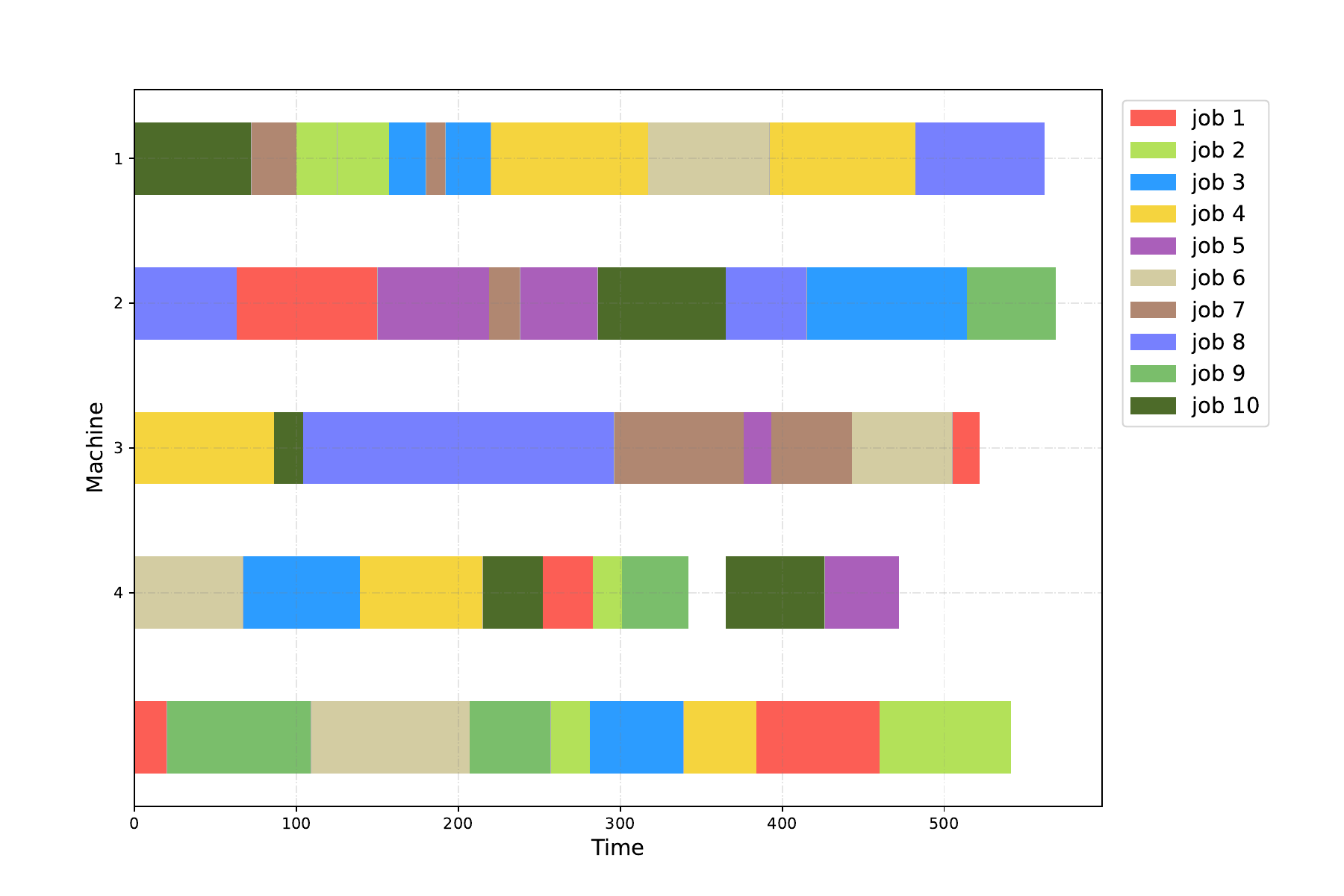}}

\caption{Hurink Rdata 5.}  
\label{fig:HurinkRdata5}
\end{figure*}

\begin{figure*}[h]
\centering
\subfloat[HGNN  Makespan:2198  Gap:2.85\%]{\includegraphics[width=0.495\textwidth]{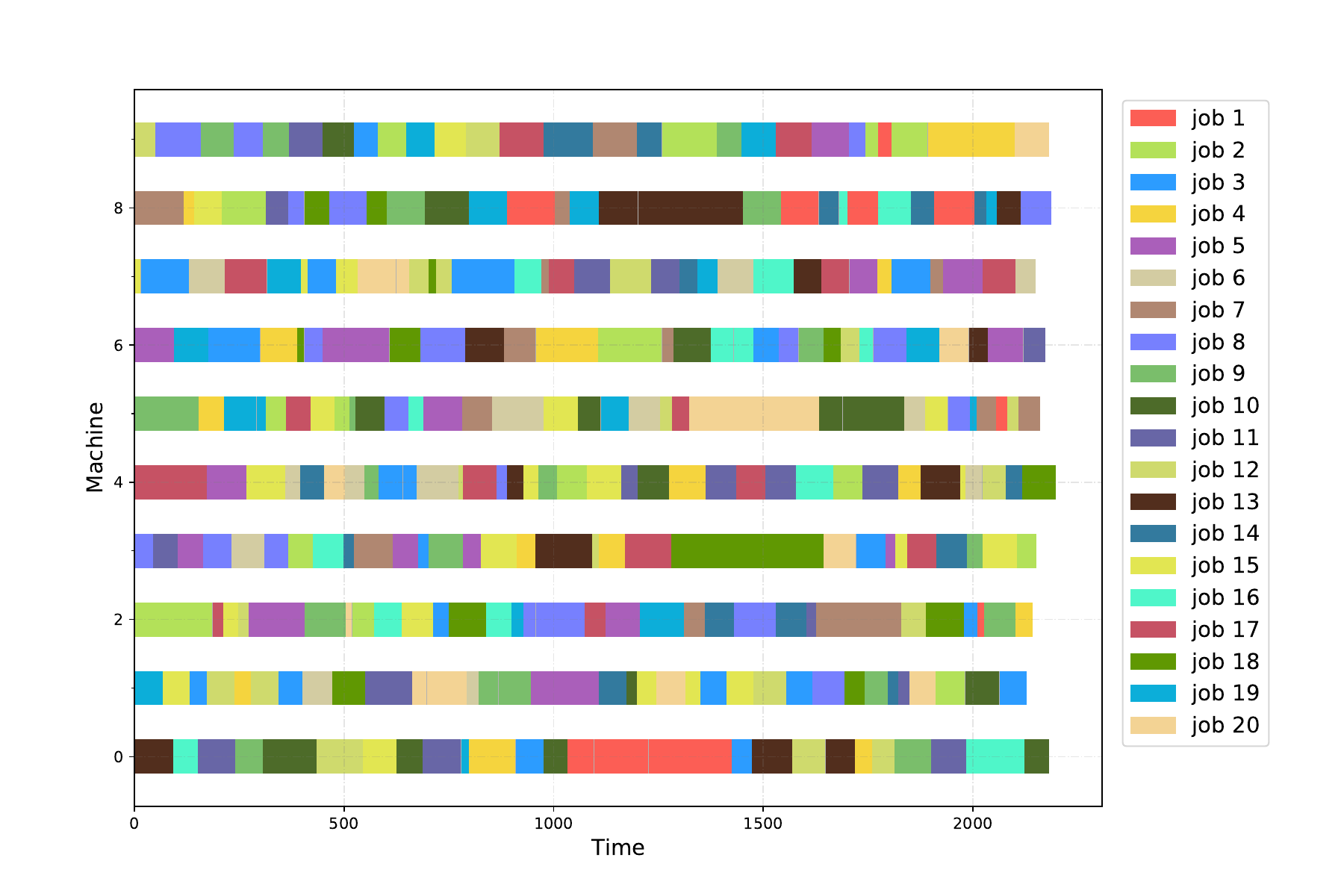}}\hspace{0.01cm}  
\subfloat[DAN  Makespan:2172  Gap:1.64\%]{\includegraphics[width=0.495\textwidth]{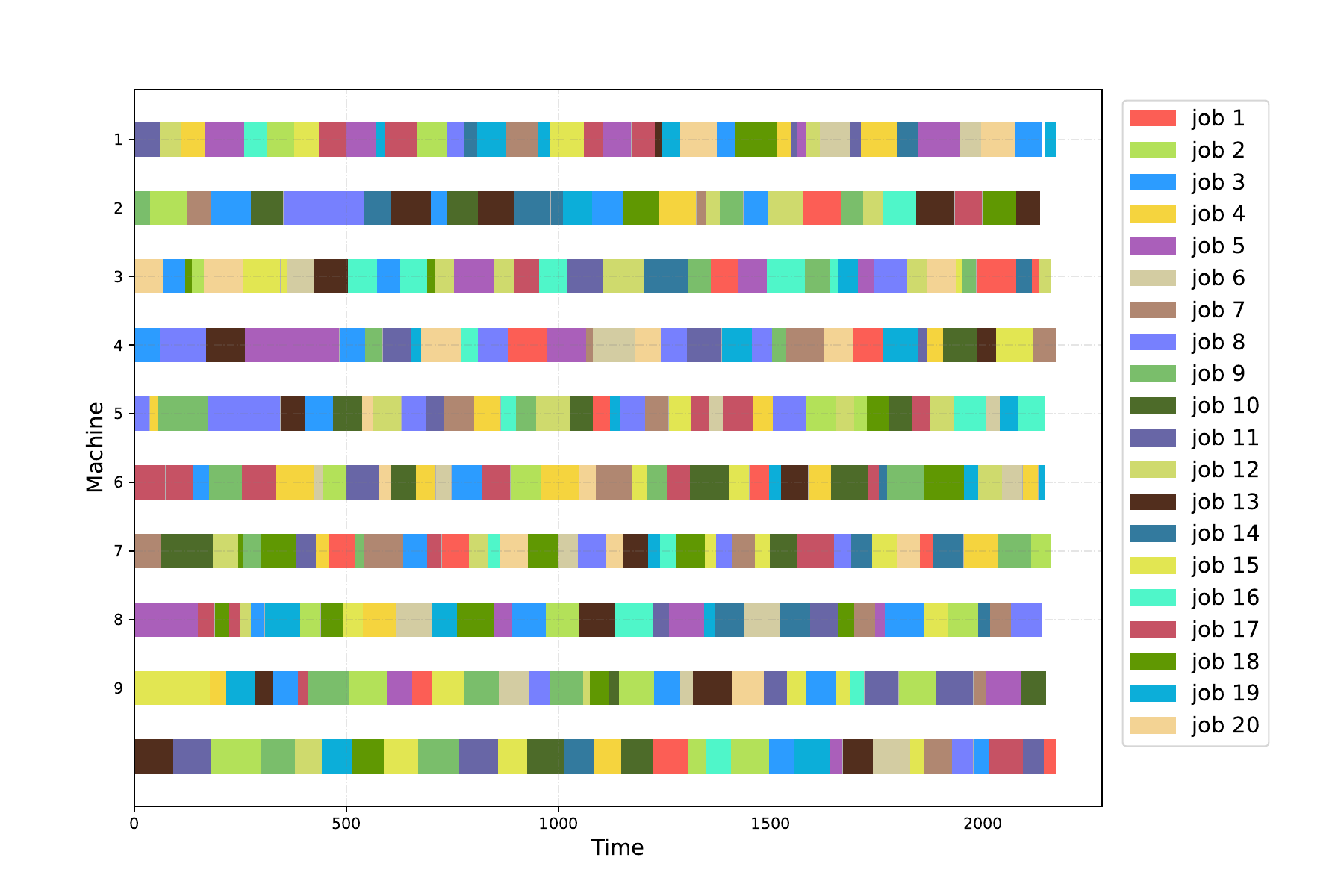}}
 \\  
\subfloat[DME  Makespan:2158  Gap:0.98\%]{\includegraphics[width=0.495\textwidth]{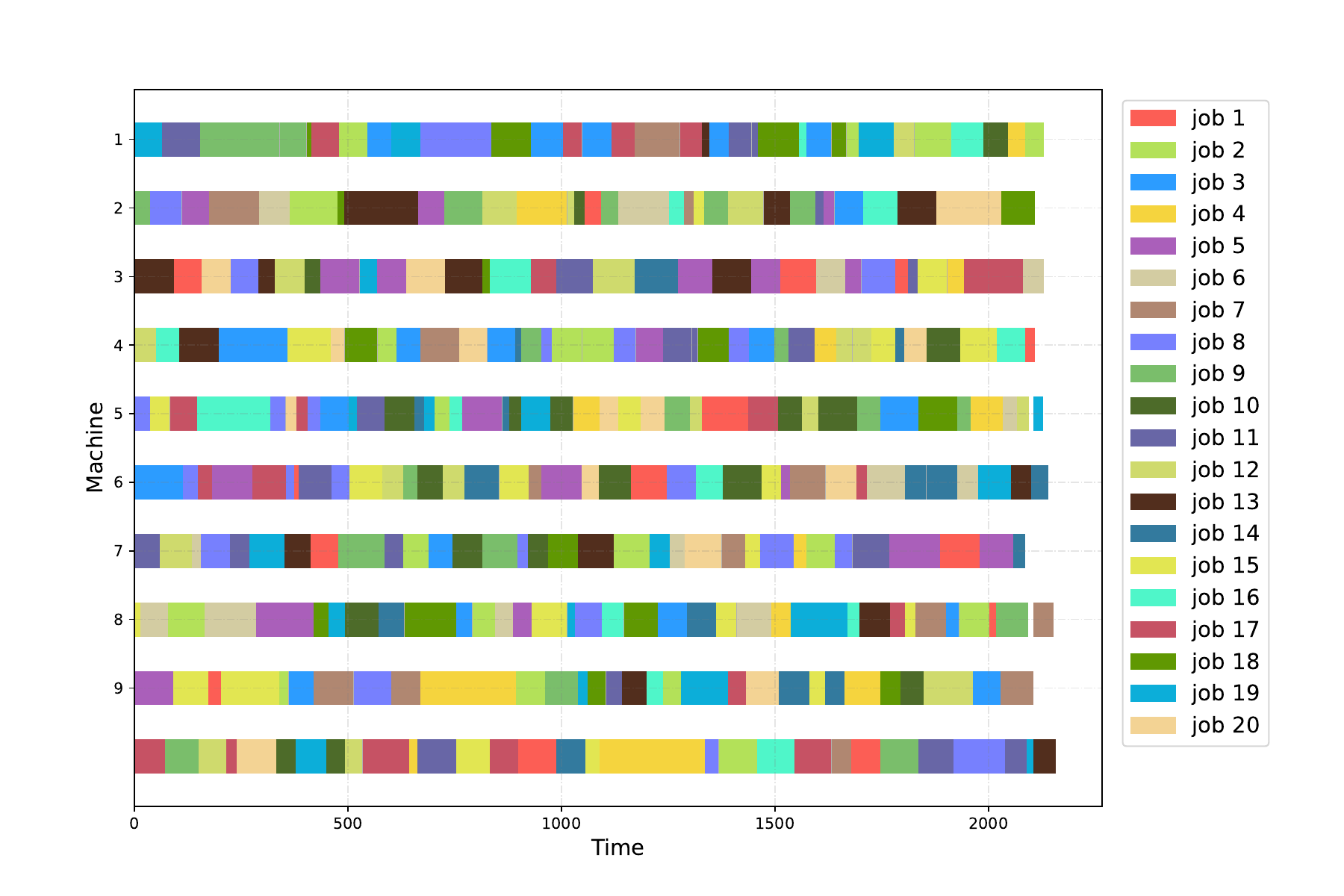}}\hspace{0.01cm}
\subfloat[M-CA  Makespan:2151  Gap:0.66\%]{\includegraphics[width=0.495\textwidth]{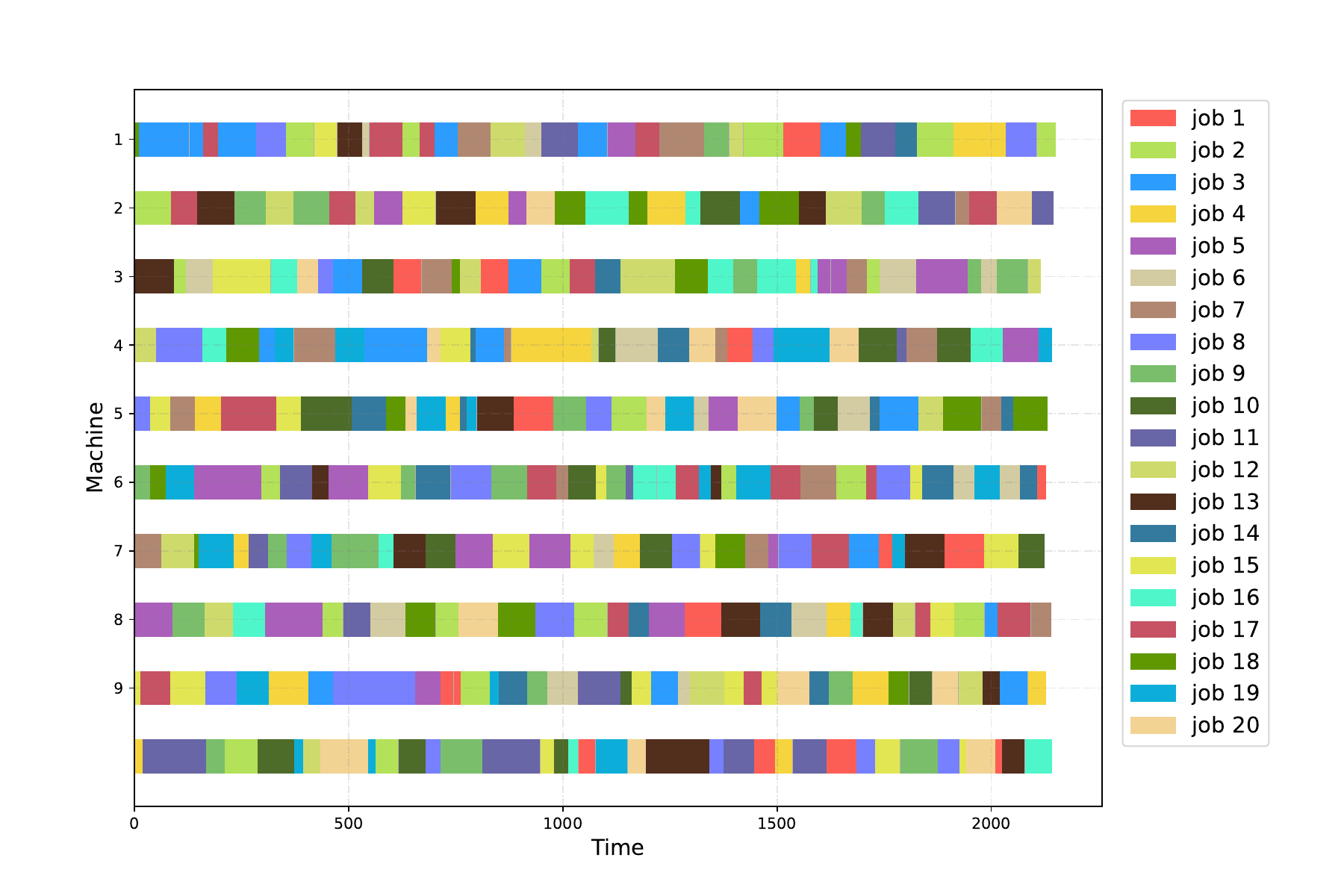}}

\caption{Dauzere 18a.}  
\label{fig:dauzere_18a}
\end{figure*}

\begin{figure*}[h]
\centering
\subfloat[L2D  Makespan:2595  Gap:29.43\%]{\includegraphics[width=0.495\textwidth]{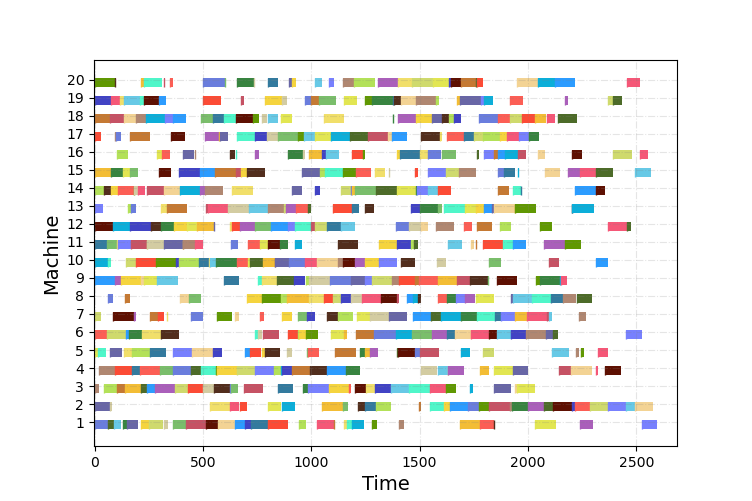}}\hspace{0.01cm}  
\subfloat[DAN  Makespan:2509  Gap:25.14\%]{\includegraphics[width=0.495\textwidth]{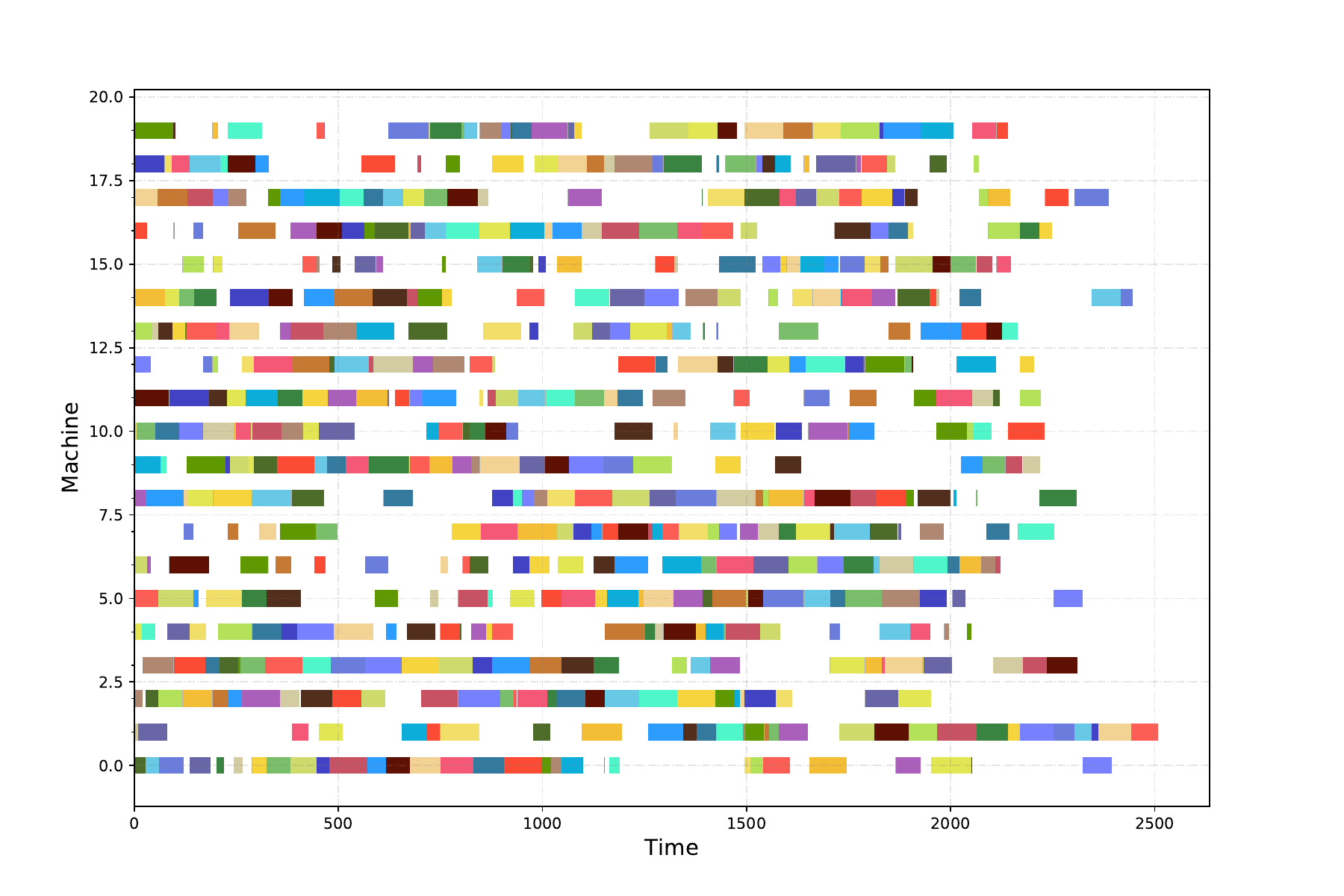}}
 \\  
\subfloat[DME  Makespan:2469  Gap:23.14\%]{\includegraphics[width=0.495\textwidth]{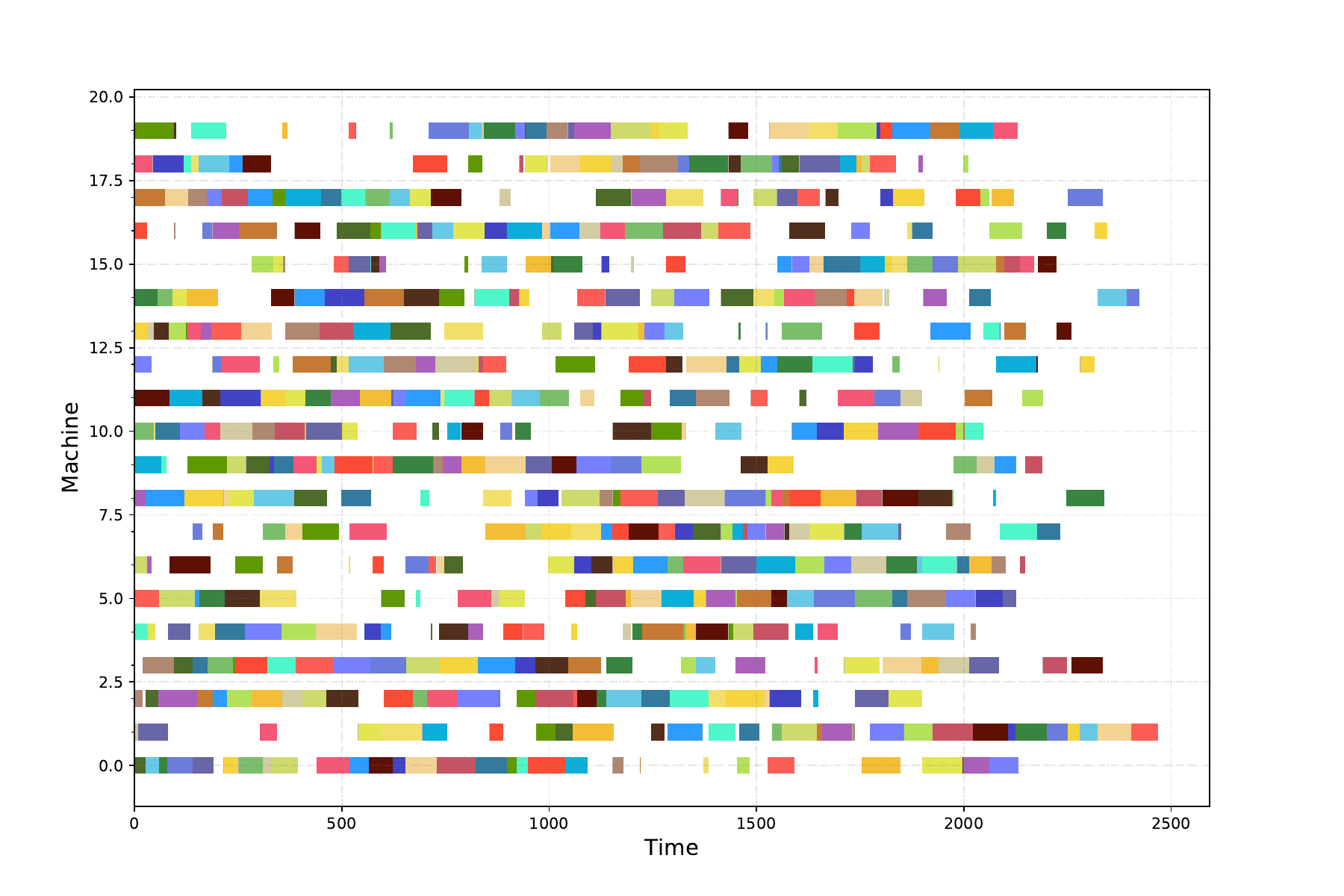}}\hspace{0.01cm}
\subfloat[M-CA  Makespan:2454  Gap:22.39\%]{\includegraphics[width=0.495\textwidth]{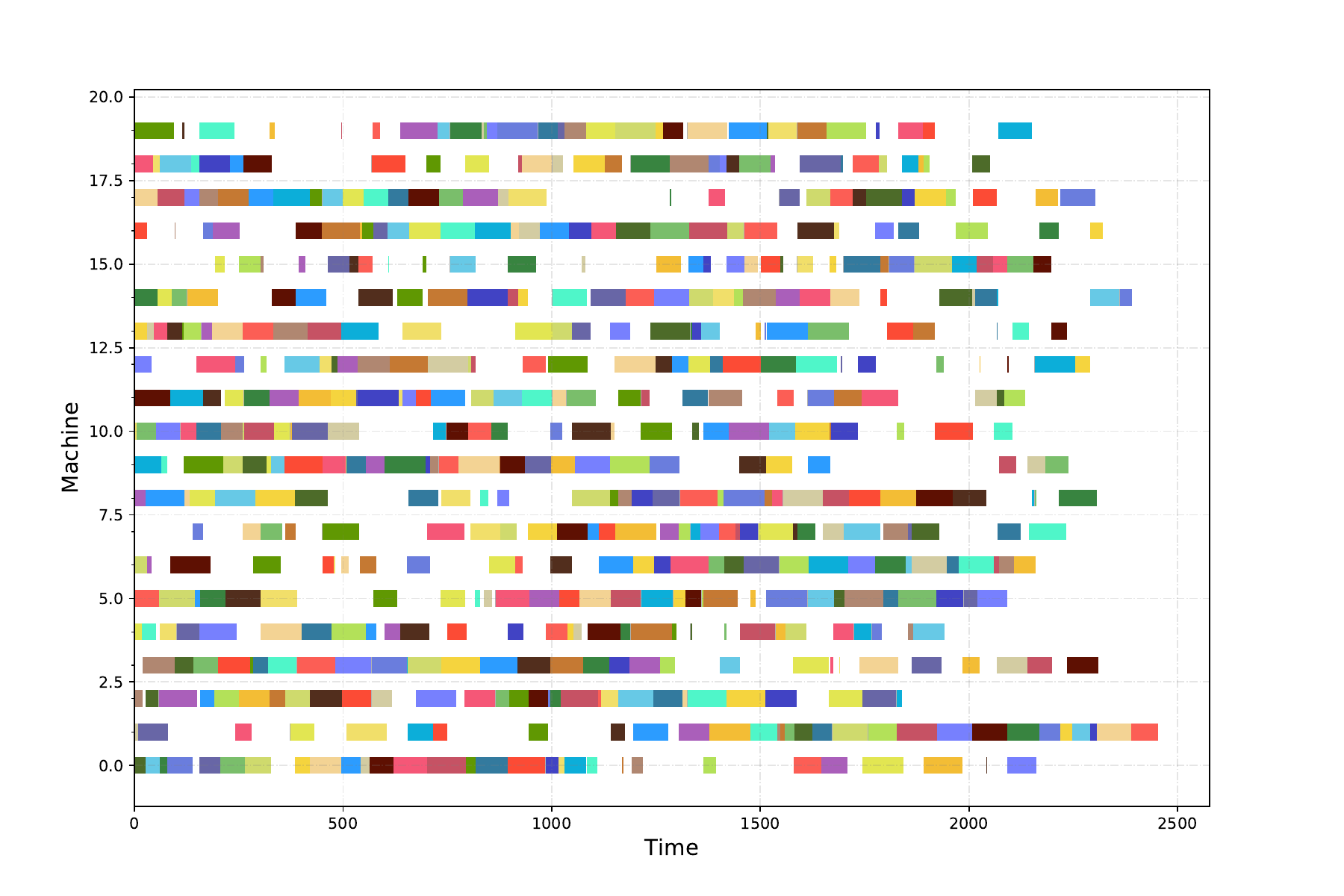}}

\caption{Taillard Ta41.}  
\label{fig:tai41}
\end{figure*}

We use Gantt charts to visualize the solutions for some of the instances in FJSP benchmarks. In Figures \ref{fig:BrandimarteMk10}, \ref{fig:HurinkRdata5}, and \ref{fig:dauzere_18a}, we compare the solutions obtained by HGNN, DAN, DME, and M-CA with the greedy strategy. 
In Figure \ref{fig:tai41}, we visualize a JSSP instance from Taillard's benchmark of size $30\times20$.

\end{document}